\documentclass[10pt, letter, onecolumn]{arxiv}

\usepackage{kantlipsum, lipsum}
\usepackage{dm-colors}
\usepackage{amsmath}
\usepackage{verbatim}
\usepackage{multirow}
\usepackage{array}
\usepackage{longtable}
\usepackage{pdflscape}
\usepackage{scalerel}
\usepackage{booktabs}
\usepackage{enumitem}
\usepackage{xspace}
\usepackage{bm}
\usepackage{bbm}
\usepackage{mathtools}
\usepackage{soul}
\usepackage{makecell}

\usepackage{graphicx}
\usepackage{amssymb}
\usepackage{colortbl}
\usepackage{lineno}
\usepackage{csquotes}
\usepackage{setspace}
\usepackage{colortbl}
\usepackage{bbding}
\usepackage{threeparttable}
\usepackage{tabularx,ragged2e}
\usepackage{placeins}
\usepackage{bbding}
\usepackage{wrapfig}

\definecolor{darkpastelgreen}{rgb}{0.13, 0.55, 0.13}

\definecolor{darkpastelred}{rgb}{0.55, 0.13, 0.13}

\definecolor{mygray}{rgb}{0.85, 0.85, 0.85}
\usepackage{xcolor}

\usepackage[hang,flushmargin]{footmisc}

\usepackage{nameref}
\usepackage{varioref}
\usepackage{amssymb}
\usepackage{pifont}
\usepackage{rotating}

\usepackage{tabularx}
\usepackage[pagebackref=false,breaklinks=false,
            colorlinks=true,bookmarks=true,citecolor=ourdarkblue,
            urlcolor=ourdarkblue,linkcolor=ourdarkblue]{hyperref}
\usepackage[noabbrev,capitalize]{cleveref}
\usepackage{etoc}

\usepackage{algorithm}
\usepackage{algpseudocode}
\usepackage{fvextra}

\usepackage{amsmath}
\usepackage{seqsplit}
\usepackage{thmtools}
\usepackage[most]{tcolorbox}
\usepackage{lmodern}
\declaretheoremstyle[
    spaceabove=6pt, spacebelow=6pt,
    headfont=\bfseries, headpunct={.}, headformat={\NAME\ \NUMBER},
    bodyfont=\normalfont,
    postheadspace=0.5em
]{promptstyle}

\tcolorboxenvironment{prompt}{
    colback=gray!10!,
    colframe=gray!75!,
    fonttitle=\bfseries,
    title=Prompt,
    boxrule=0.5pt,
    sharp corners,
    breakable
}

\graphicspath{{figures/}}

\newcommand{\rev}{\color{black}}

\newcommand{\benchname}{MedSP1000\xspace} 

\title{\Large{Evaluating Large Language Models in Dynamic Clinical Decision-Making with Standardized Patient Cases}}

\author[1,2, $\ast$]{Cheng Liang}
\author[1,2, $\ast$]{Pengcheng Qiu}
\author[1,2]{Ya Zhang}
\author[1,2,$\dag$]{\\ \vspace{0.1cm}Yanfeng Wang}
\author[1,$\dag$]{Chaoyi Wu} 
\author[1,2,$\dag$]{Weidi Xie}

\affil[1]{\normalsize Shanghai Jiao Tong University, Shanghai, China \authorcr \vspace{0.1cm}}

\affil[2]{\normalsize Shanghai Artificial Intelligence Laboratory, Shanghai, China  \authorcr \vspace{0.1cm}
}
\affil[$\ast$]{\normalsize Equal contributions\hspace{1cm}}
\affil[$\dag$]{\normalsize Corresponding author
\authorcr Cheng Liang: liangcheng1026@sjtu.edu.cn; Pengcheng Qiu: henrychur@sjtu.edu.cn \authorcr Chaoyi Wu: wtzxxxwcy02@sjtu.edu.cn; Weidi Xie: weidi@sjtu.edu.cn }

\begin{document}


\begin{abstract}
Large language models (LLMs) are increasingly proposed as clinical agents, yet static, single-turn benchmarks cannot capture how a model dynamically delivers care across an encounter: gathering information, planning treatment, and adapting longitudinal management across successive patient states.
Medical education has long addressed an analogous challenge through standardized patients (SPs): trained actors who consistently portray clinical cases to enable learners to practice in safe simulations that closely mimic real-world clinical dynamics while supporting objective, quantifiable assessment through standardized scripts.
Here we introduce \benchname, an SP-derived interactive benchmark for clinical-agent evaluation, including 1,638 SP cases with 24,602 trajectory-level peer-reviewed rubrics. \benchname converts peer-reviewed SP teaching cases into executable scenarios with defined SP case scripts, clinical environment contexts, and human-validated structured rubric. In each simulation evaluation run, a clinical agent interacts in closed loop with a patient agent and an environment controller, and its behaviour is scored throughout the encounter against expert criteria specified in the original materials. 

Applying \benchname to a range of general-purpose and medically specialized LLMs, we find that performance on static benchmarks does not reliably translate to such educational scenarios. The best-performing model, GPT-5.5, completes only 60.4\% of expert-defined rubric items, whereas the strongest medically specialized model reaches 40.0\%; increasing test-time compute produces no measurable gain. These results suggest that current LLMs, including agentic systems tuned for medicine, are not yet reliable enough to be safely integrated into actual clinical practice. More broadly, \benchname shows how process-level, SP-style evaluation can reveal clinically relevant failure modes that single-turn benchmarks miss.

\end{abstract}

\maketitle

\renewcommand\labelitemi{{\boldmath$\circ$}}

\section{Introduction}

Large language models (LLMs) have achieved strong performance on medical examinations~\cite{singhal2023large, singhal2025toward, mcduff2025towards} and clinical decision-support benchmarks~\cite{wu2025towards,sandmann2025benchmark, gaber2025evaluating,qiu2025quantifying}. However, most evaluations still focus on fixed medical question answering rather than the ability to function as interactive clinical agents, a capability critical to real-world practice~\cite{moritz2025coordinated, freyer2025overcoming,lin2026framework,teo2025generative}. Clinical care unfolds over time: clinicians must gather incomplete information, choose examinations and tests, interpret evolving results, initiate interventions, monitor responses, communicate with patients and revise management plans as new evidence emerges. Strong performance on static questions therefore does not establish competence in the sequential reasoning and action required for real clinical encounters.

Current evaluation paradigms capture only part of this process. These benchmarks rely heavily on single-turn question-answering datasets~\cite{jin2021disease, pal2022medmcqa, jin-etal-2019-pubmedqa,zuo2025medxpertqa,healthbench,helm,hager2024evaluation}, which flatten clinical practice into isolated prompts and final answers. Recent efforts have introduced more interactive settings~\cite{johri2025evaluation, agentclinic,tu2025towards,agenthospital, qiu2025evolving}, but many remain focused on diagnostic history-taking or online consultation. As a result, they provide limited assessment of broader clinical behaviours such as interventional decision-making, longitudinal treatment planning, monitoring for complications and adapting management across successive patient states.

Medical education offers a mature framework for evaluating such skills. Standardized patients (SPs) are trained actors who consistently portray clinical cases, allowing learners to practise and be assessed in realistic encounters without exposing real patients to avoidable risk. In objective structured clinical examinations~(OSCEs), trainees interact with SPs and are evaluated by faculty using predefined rubrics that enable objective, quantitative scoring of not only diagnostic accuracy, but also communication, professionalism, clinical reasoning, and management. This makes SP cases a natural foundation for evaluating LLMs as clinical agents: they are interactive by design, standardized across examinee clinicians and grounded in expert-authored assessment criteria.

\begin{figure}[t]
    \centering
    \includegraphics[width=\textwidth]{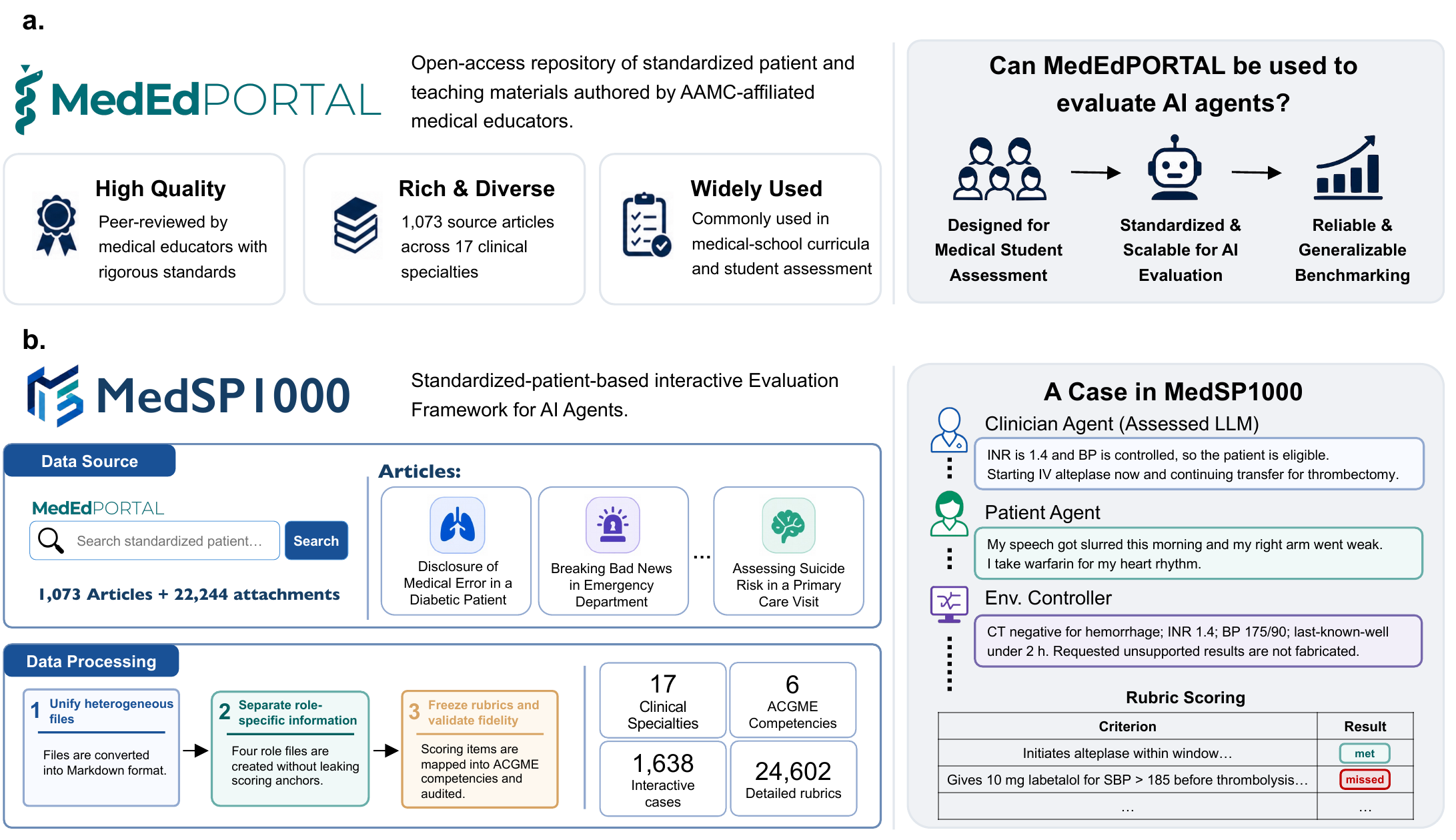}
    \caption{
    \textbf{Overview of \benchname.}
    \textbf{a,} \benchname adapts peer-reviewed standardized-patient and simulation cases from MedEdPORTAL into closed-loop evaluations for clinical agents. A target LLM acts as the clinician, interacts with patient and environment agents, and is scored against expert rubrics derived from the original teaching materials.
    \textbf{b,} heterogeneous source files are parsed, partitioned into role-specific materials and audited for leakage and consistency. The resulting benchmark contains 1{,}638 interactive scenarios spanning 17 specialties and the six ACGME core competencies.}
    \label{fig:teaser}
\end{figure}

Here we introduce \benchname, an interactive benchmark for clinical LLM evaluation built from 1{,}638 SP cases with 24{,}602 trajectory-level rubrics curated from peer-reviewed medical education materials in MedEdPORTAL. As shown in Figure~\ref{fig:teaser}, the benchmark spans 17 clinical specialties and evaluates model behaviour across the six core competencies of the Accreditation Council for Graduate Medical Education (ACGME): patient care, medical knowledge, practice-based learning and improvement, systems-based practice, professionalism, and interpersonal and communication skills~\cite{stanford_acgme_2026}. Each case is converted into an executable scenario with a clinical context, role-specific materials, predefined patient and environment responses, and tailored scoring rubrics derived from the original teaching materials.  

To construct \benchname at scale, we developed a three-phase curation pipeline that converts heterogeneous SP teaching documents into structured materials. 
The pipeline first consolidates raw source attachments into a unified Markdown format, then partitions it into four role-specific packets containing only the information available to that role:
a scenario initialization, a patient script, environment-controller materials and a scoring rubric. A final self-reflective audit step checks for leakage between roles and verifies consistency with the source materials. 
To assess the reliability of this automated process, twelve clinicians reviewed 100 constructed SP cases, with each scenario independently scored by two annotators for source interpretation, structural completeness, clinical fidelity and simulation feasibility. Each extracted element was linked to its source passage to enable traceable review. 
Clinicians rated the constructed cases highly across all four aspects (mean 4.78 on a 5-point scale), with close inter-annotator agreement (mean difference 0.41), supporting their fidelity to the original teaching intent.
We instantiate a closed-loop multi-agent evaluation framework. In each encounter, the model under evaluation acts as the examinee clinician. It starts with scenario initialization input, interacts with a patient agent that handles communication and symptom presentation, and with an environment controller that resolves physical examinations, diagnostic tests, procedures, and clinical state transitions. The environment controller advances the clinical scenario only when permitted by the source materials, ensuring that different models face the same clinical trajectory under standardized conditions. The completed trajectory is then scored by an evalutor agent item by item against the expert rubrics, enabling process-level evaluation of what the model did, omitted or did at the wrong time.

We systematically evaluate a representative set of seven LLMs on \benchname, including frontier proprietary models (GPT-5.5, Claude-Opus-4.7, Gemini-3.1-Pro), leading open-source general-purpose models (DeepSeek-V4-Pro, Qwen-3.5), and medical-domain models (MedGemma, Baichuan-M3). The results reveal a substantial gap between static medical competence and interactive clinical performance. The strongest model, GPT-5.5, completes only 60.4\% of expert-defined rubric items, whereas the best medical-specialized model reaches 40.0\%, trailing GPT-5.5 by 20.4 percentage points. Across models, performance is particularly weak in Practice-Based Learning and Improvement, the ACGME competency associated with self-reflection, error recognition, and iterative improvement, where no model exceeds 30\% completion. Additional test-time compute does not close this gap: on the subset used for this analysis, applying Best-of-\(N\) sampling and a MDT-style multi-agent strategy to GPT-5.5 changes its macro completion rate from 67.1\% to 67.8\% and 68.0\%, respectively. Together, these findings show that current LLMs, including models specialized for medicine, remain far from reliable interactive clinical agents, and that SP-derived process-level evaluation reveals clinically important failure modes that static benchmarks miss.

\section{Results}

In this section, we first introduce the \benchname and its evaluation protocol. We then present the experimental performance of each model across simulated clinical cases, followed by a synthesis of the key findings and capability patterns revealed by these results. 

\subsection{Introduction of \benchname}

\subsubsection{Standardized Patient Construction}

The simulation cases in \benchname are curated from MedEdPORTAL educational resources, primarily including articles related to the SPs. The instructional SP materials already specify task objectives, case setups, role assignments, clinical progression and grading points, which enable us to construct faithful simulated clinical environments.

Based on these original materials, we ultimately constructed an evaluation set comprising \textbf{1,638} executable cases. Rather than simply flattening all the information from the original attachments and grading sheets, we extracted, filtered, and structured the key content into four role-specific packets: a scenario initialization, a patient script, environment-controller materials, and a scoring rubric. Based on this split, we build up a dynamic evaluation framework. Specifically, during evaluation runtime, for each SP case, the model under evaluation is placed in the role of the clinician and engages in closed-loop interactions with a patient agent and an environment controller. It first receives only the ``scenario initialization'' packet, which contains the general scenario description and initial chief complaint. It needs to progressively interact with other agents to obtain further necessary information. Correspondingly, the patient and environment-controller agents receive the ``patient script'' and ``environment-controller materials'' packets, respectively, which detail how they should respond to the assessed clinician agent's queries. Lastly, the evaluator receives the ``scoring rubric'' packet to make binary judgments on whether the multi-turn trajectory meets each grading criterion. In this way, the scenario background, role information, dynamic events, and grading references, which originally dispersed across multiple files, are reorganized into executable structured materials to support an effective multi-agent SP simulation. For more details on the data curation of \benchname and the setup of the evaluation framework, please refer to Section~\ref{sec:data_curation} and Section~\ref{sec:evaluation_framework}.

\begin{figure}[tbp]
    \centering
    \includegraphics[width=1\linewidth]{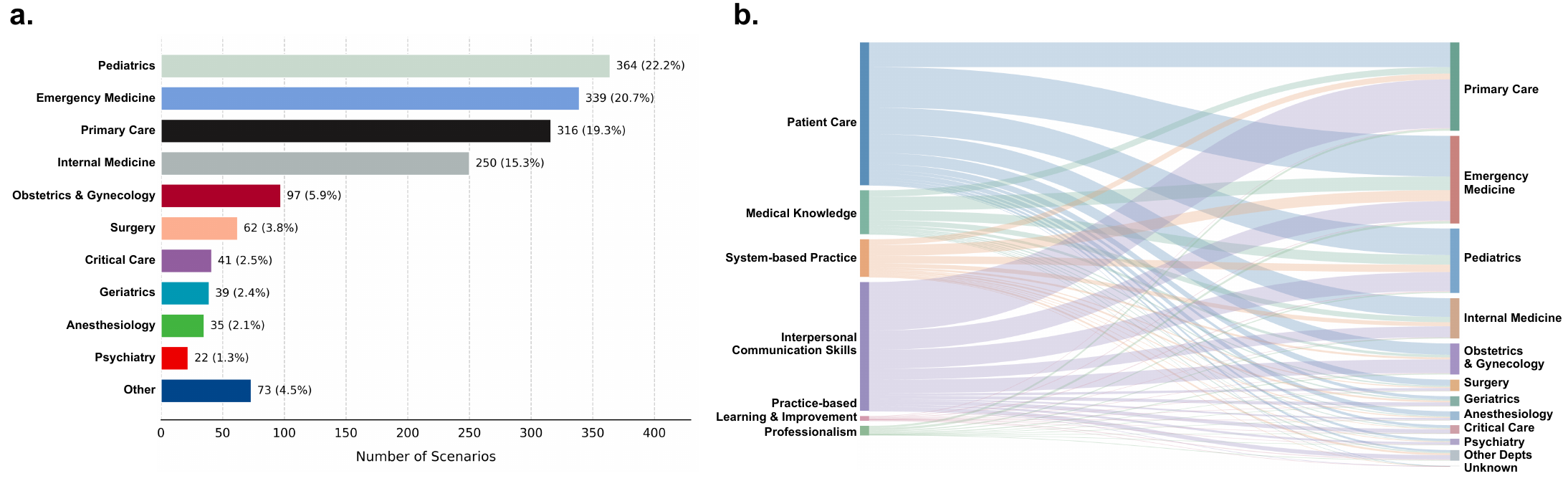}
    \caption{\textbf{\benchname cases composition.}
    \textbf{a.} Distribution of cases across clinical specialties ($N=1{,}638$).
    \textbf{b.} Sankey diagram of rubric items linking the six ACGME competencies to the clinical specialties.}
    \label{fig:result_data_dist}
\end{figure}

The final \benchname span a broad range of clinical specialties, whose distribution is shown in Figure~\ref{fig:result_data_dist}a. Each case is assigned to a medical specialty following the taxonomy built in HealthBench Professional~\cite{healthbench-pro} and each rubric item is assigned to one of the six ACGME core competencies. The Sankey diagram in Figure~\ref{fig:result_data_dist}b further shows how the SP cases link those competencies to the individual specialties, characterizing both the range of scoring requirements the benchmark covers and how they distribute across competencies and specialties.

\subsubsection{Human Validation of \benchname}
\label{sec:human_validation}

We further ensure the reliability of \benchname through clinician validation by collaborating with twelve clinicians (mean 8 years of medical training) from different affiliated hospitals of Shanghai Jiao Tong University, confirming that both the data-construction pipeline and the evaluation framework are trustworthy.

\noindent\textbf{Data construction quality.} We conducted a systematic human validation on the final structured SP case packets. The clinicians reviewed 100 automatically constructed cases, with each case independently scored by two annotators to enable cross-annotation and inter-rater agreement analysis. 
Using a 5-point scale, they assessed four aspects of construction quality: 
\textit{(i)} file reading and understanding: whether the pipeline correctly interpreted the source instructional materials, 
\textit{(ii)} output structure: whether the resulting role-specific packets were structurally complete and well organized, 
\textit{(iii)} source consistency: whether the constructed materials remained strictly faithful to the source in clinical content and scoring criteria, 
and \textit{(iv)} simulation plausibility: whether the resulting case was suitable for stable text-based simulation. 
Potential information leakage across role-specific packets was also examined as part of the structural and source-fidelity review, to confirm that each agent received only role-appropriate content.

\begin{figure}[tbp]
    \centering
    \includegraphics[width=1\linewidth]{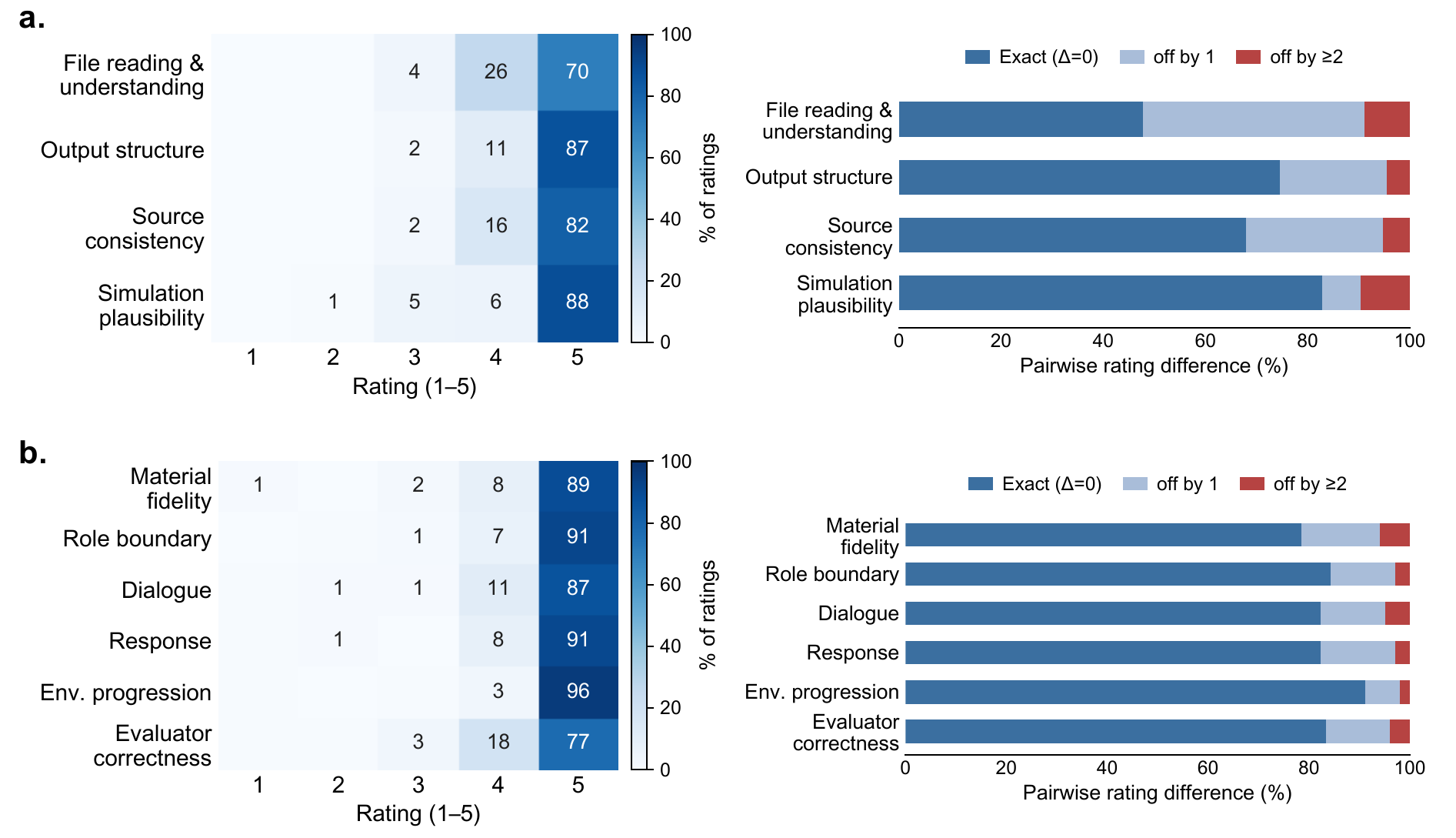}
    \caption{\textbf{Clinician evaluation of \benchname by 12 clinicians, with each case independently scored by two annotators.}
    \textbf{a.} Data-construction quality, assessed across four dimensions. The heatmap shows the rating distribution, and the stacked bar shows the pairwise inter-annotator difference.
    \textbf{b.} Runtime simulation quality, assessed across six dimensions, presented in the same format.}
    \label{fig:result_data_human_eval}
\end{figure}

As shown in Figure~\ref{fig:result_data_human_eval}a, ratings are high across all four aspects, with mean scores of 4.66, 4.85, 4.80, and 4.81, respectively. The two annotators are also highly consistent, differing by only 0.41 points on average, indicating that the human validation is robust. Together, these results demonstrate that the pipeline reliably produces evaluation-ready cases while preserving fidelity to the original teaching materials. Notably, during the validation process, the clinicians also corrected any identified errors, yielding a human-verified clean subset \textbf{MedSP1000-Verified}, including 100 cases. We report the per-model rubric completion rate on this subset in Supplementary Table~\ref{tab:subset_performance}.

\noindent\textbf{Evaluation framework quality.} Beyond construction quality, we assessed how faithfully the framework executes at runtime, using the simulation runs produced from the same 100 validated SP cases. Twelve clinicians reviewed these 100 completed runs, with each run also independently scored by two annotators to enable cross-annotation and inter-rater agreement analysis. Similarly, using a 5-level scale, they assessed six aspects of runtime simulation fidelity: \textit{(i)} material fidelity: whether the agents' outputs remained faithful to the source materials, \textit{(ii)} role boundary: whether role boundaries were correctly preserved throughout the interaction, \textit{(iii)} dialogue: whether the dialogue was clinically coherent, \textit{(iv)} response: whether agent responses were appropriate to the evolving scenario state, \textit{(v)} environment progression: whether environmental progression was correctly triggered and advanced, and \textit{(vi)} evaluator correctness: whether the evaluator agent's per-rubric judgements agreed with human adjudication. 
The first four aspects are targeted at the patient and environment controller agents, while the last one is for the evaluator.

As shown in Figure~\ref{fig:result_data_human_eval}b, ratings across all six dimensions were concentrated in the upper end of the scale, with 97.7\% of all ratings falling in the 4-5 range. Cross-annotation agreement was likewise high: exact agreement accounted for 83.3\% of paired ratings overall, and 96.2\% of paired ratings differed by no more than 1 point. Together, these results indicate that the simulation framework can execute the role-specific packets reliably during interactive evaluation runtime.

\subsubsection{Evaluation Metric}

Inspired by the HealthBench~\cite{healthbench}, we implement the rubric-based evaluation metric. Each case carries a set of rubrics which are organized along the six core competencies defined by the Accreditation Council for Graduate Medical Education (ACGME): patient care (PC), medical knowledge (MK), systems-based practice (SBP), interpersonal and communication skills (ICS), practice-based learning and improvement (PBLI), and professionalism (PROF). After a scenario run concludes, an evaluator agent judges each rubric item as completed or not based on the full interaction trajectory. We report the case-level \emph{rubric completion rate} (the fraction of rubric items completed in a run) as the primary metric, which is straightforwardly calculated by averaging the completion rates across all cases. Additionally, we also adopt specialty-level and competency-level rubric completion rate. For the specialty-level, we first calculate the rate per case and then aggregate them through averaging. Meanwhile, for the competency-level, considering that competencies are defined at the rubric item level instead of the case level, we directly calculate the completion rate across all rubric items belonging to the same competency. Full definitions are given in Section~\ref{sec:method_eval_metrics}. Unless otherwise specified, in following the term \emph{rubric completion rate} refers to the case-level metric by default.

\begin{figure}[tbp]
    \centering
    \includegraphics[width=1\linewidth]{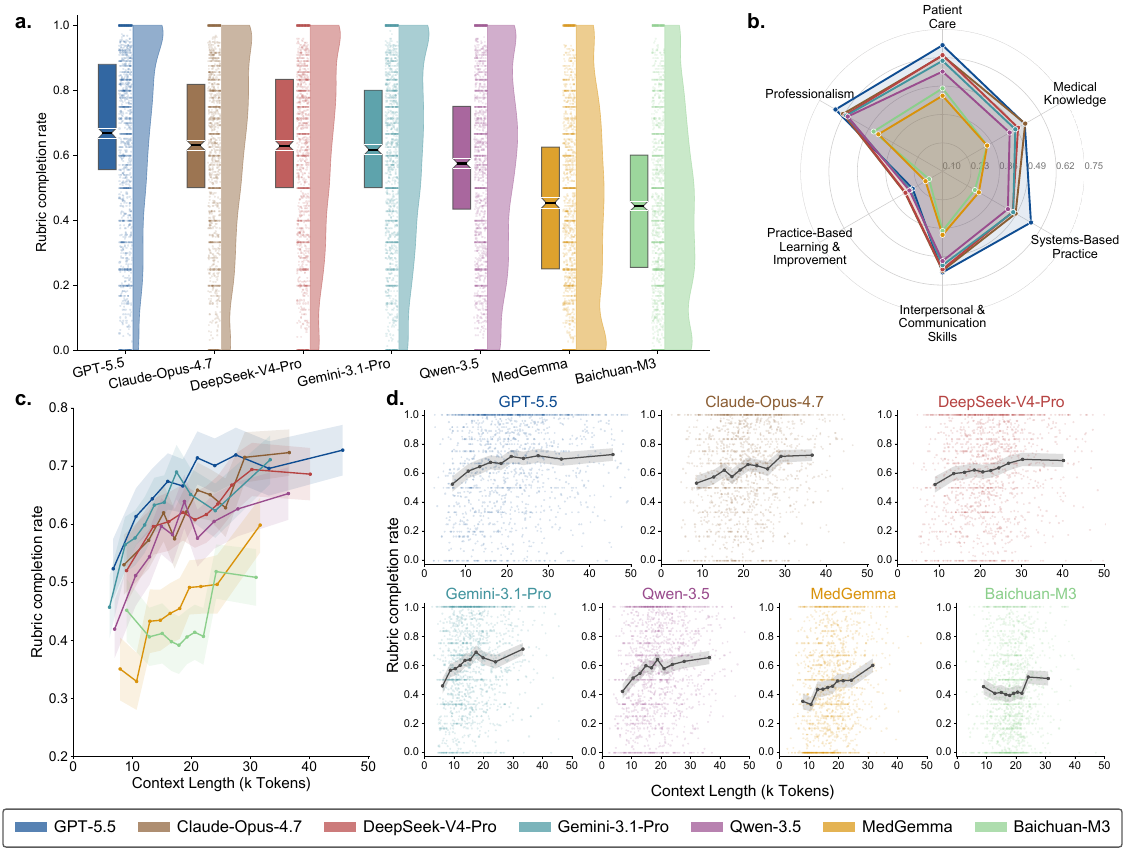}
    \caption{\textbf{Model performance on \benchname.}
    \textbf{a.} Overall performance. The raincloud plot shows the per-case macro rubric completion rate. The summary bar (left) carries five horizontal markers, from bottom to top, the 33rd percentile, the lower bound of the 95\% CI of the mean, the mean (black line), the upper bound of the 95\% CI, and the 66th percentile. The points (middle) are the individual per-case completion rates, and the curve (right) is a kernel density estimate of their distribution.
    \textbf{b.} Per-competency capability results across the six ACGME competencies, shown as a radar plot of the micro rubric completion rate for each evaluated model.
    \textbf{c.} Relationship between macro rubric completion rate and assessed model input length; each line is one model and the shaded band is the 95\% CI.
    \textbf{d.} The same relationship decomposed by model: in each panel, points are individual cases (completion rate versus assessed model's input length), the black line is the binned trend, and the shaded band is the 95\% CI.}
    \label{fig:results}
\end{figure}

\subsection{Evaluation Results across Core Competencies}
\label{sec:overall_performance}

\begin{table}[!t]
    \renewcommand{\arraystretch}{1.3}
    \footnotesize
    \centering
    \caption{Overall rubric completion rate and per-competency capability profile on \benchname. All values are percentages. PC, patient care; MK, medical knowledge; SBP, systems-based practice; ICS, interpersonal and communication skills; PBLI, practice-based learning and improvement; PROF, professionalism. \emph{Micro} pools all rubric items across cases before averaging; \emph{Macro} is the competency-macro score, computed as the unweighted average of the six per-competency scores. Size denotes total parameters; Context denotes the maximum supported context-window length in tokens (K, thousand; M, million). ``--'' indicates information not publicly disclosed. \textbf{Bold} marks the best value in each column. In each cell, the top line is the point estimate and the bottom line is its 95\% confidence interval (bootstrap over cases).}
    \label{tab:overall_performance}
    \resizebox{\textwidth}{!}{
    \begin{tabular}{l|ccc|cccccc|cc}
    \toprule
    Model & Size & Context & Date & PC & MK & SBP & ICS & PBLI & PROF & Micro & Macro \\
    \midrule
    \rowcolor{mygray} \multicolumn{12}{c}{Closed-source LLMs} \\
    \midrule
    GPT-5.5 & -- & 1M & Apr 2026 & \makecell{\textbf{67.7}\\{\scriptsize (65.3--70.1)}} & \makecell{53.4\\{\scriptsize (49.7--57.3)}} & \makecell{\textbf{56.7}\\{\scriptsize (52.9--60.5)}} & \makecell{\textbf{56.2}\\{\scriptsize (53.6--58.9)}} & \makecell{25.8\\{\scriptsize (18.9--33.7)}} & \makecell{\textbf{66.6}\\{\scriptsize (62.1--71.0)}} & \makecell{\textbf{60.4}\\{\scriptsize (58.7--62.3)}} & \makecell{\textbf{54.4}\\{\scriptsize (52.3--56.5)}} \\
    Claude-Opus-4.7 & -- & 1M & Apr 2026 & \makecell{62.9\\{\scriptsize (60.6--65.3)}} & \makecell{\textbf{53.7}\\{\scriptsize (49.7--57.3)}} & \makecell{48.6\\{\scriptsize (44.9--52.5)}} & \makecell{55.4\\{\scriptsize (53.0--58.1)}} & \makecell{\textbf{29.8}\\{\scriptsize (21.4--38.9)}} & \makecell{62.5\\{\scriptsize (57.6--66.9)}} & \makecell{57.4\\{\scriptsize (55.7--59.1)}} & \makecell{52.1\\{\scriptsize (49.8--54.5)}} \\
    Gemini-3.1-Pro & -- & 1M & Feb 2026 & \makecell{60.6\\{\scriptsize (58.0--63.3)}} & \makecell{48.4\\{\scriptsize (44.1--52.4)}} & \makecell{47.3\\{\scriptsize (43.8--50.9)}} & \makecell{52.9\\{\scriptsize (50.6--55.3)}} & \makecell{27.0\\{\scriptsize (20.1--34.4)}} & \makecell{60.9\\{\scriptsize (56.3--65.6)}} & \makecell{54.7\\{\scriptsize (53.0--56.6)}} & \makecell{49.5\\{\scriptsize (47.4--51.8)}} \\
    \midrule
    \rowcolor{mygray} \multicolumn{12}{c}{Open-source General LLMs} \\
    \midrule
    DeepSeek-V4-Pro & 1.6T & 1M & Apr 2026 & \makecell{63.1\\{\scriptsize (60.7--65.8)}} & \makecell{49.9\\{\scriptsize (45.8--53.8)}} & \makecell{47.1\\{\scriptsize (43.5--50.8)}} & \makecell{54.8\\{\scriptsize (52.1--57.6)}} & \makecell{29.6\\{\scriptsize (21.9--38.3)}} & \makecell{61.4\\{\scriptsize (56.8--66.6)}} & \makecell{56.6\\{\scriptsize (54.9--58.4)}} & \makecell{51.0\\{\scriptsize (48.7--53.3)}} \\
    Qwen-3.5 & 397B & 256K & Mar 2026 & \makecell{55.6\\{\scriptsize (53.4--58.1)}} & \makecell{45.5\\{\scriptsize (41.8--48.9)}} & \makecell{44.5\\{\scriptsize (41.0--48.3)}} & \makecell{51.0\\{\scriptsize (48.4--53.6)}} & \makecell{27.5\\{\scriptsize (20.4--35.3)}} & \makecell{60.1\\{\scriptsize (55.7--65.0)}} & \makecell{51.5\\{\scriptsize (49.9--53.1)}} & \makecell{47.4\\{\scriptsize (45.4--49.5)}} \\
    \midrule
    \rowcolor{mygray} \multicolumn{12}{c}{Medical LLMs} \\
    \midrule
    Baichuan-M3 & 235B & 41K & Jan 2026 & \makecell{47.9\\{\scriptsize (45.6--50.3)}} & \makecell{33.6\\{\scriptsize (29.6--37.3)}} & \makecell{27.1\\{\scriptsize (24.3--29.9)}} & \makecell{37.2\\{\scriptsize (35.2--39.6)}} & \makecell{17.1\\{\scriptsize (11.3--23.6)}} & \makecell{46.4\\{\scriptsize (41.4--51.2)}} & \makecell{40.0\\{\scriptsize (38.5--41.6)}} & \makecell{34.9\\{\scriptsize (33.0--36.9)}} \\
    MedGemma & 27B & 128K & Jul 2025 & \makecell{44.6\\{\scriptsize (42.4--47.0)}} & \makecell{33.6\\{\scriptsize (30.2--37.0)}} & \makecell{29.2\\{\scriptsize (26.4--32.0)}} & \makecell{39.1\\{\scriptsize (36.6--41.4)}} & \makecell{19.0\\{\scriptsize (12.8--26.6)}} & \makecell{43.9\\{\scriptsize (38.8--49.7)}} & \makecell{39.5\\{\scriptsize (37.7--41.1)}} & \makecell{34.9\\{\scriptsize (32.8--37.0)}} \\
    \bottomrule
    \end{tabular}
    }
\end{table}

We evaluate seven representative LLMs on \benchname, spanning three model families: closed-source frontier models~(GPT-5.5, Claude-Opus-4.7 and Gemini-3.1-Pro), open-source general-purpose models~(DeepSeek-V4-Pro and Qwen-3.5), and medical-specialized models~(Baichuan-M3 and MedGemma). Detailed descriptions of these models are provided in Section~\ref{sec:baselines}. To ensure reproducibility, all LLMs, including those deployed in our multi-agent evaluation framework and the assessed clinician one, are implemented using deterministic decoding~(temperature = 0). We report each model's rubric completion rate on \benchname as follows.

Figure~\ref{fig:results}a reports the overall rubric completion rate across different models and the Figure~\ref{fig:results}b further presents a \emph{competency-level} view via a radar plot of each model's rubric completion rate across the six ACGME core competencies. Table~\ref{tab:overall_performance} further reports the detailed competency-level scores. In the table, two complementary average metrics are reported: the \emph{micro} rate, which is computed by pooling all relevant rubric items together (weighting each rubric item equally), and the \emph{macro} rate, which is computed by averaging the six per-competency scores so that each ACGME dimension is weighted equally.

The four highest-scoring models are all general-purpose frontier models and cluster within a six-point band on the micro metric, with GPT-5.5 in the lead at 60.4\%. The competency-macro values are uniformly 3-6 points below the micro values, because micro is dominated by the two highest-volume competencies (PC and ICS), whereas macro gives equal weight to low-volume, low-scoring dimensions such as PBLI.
 
Across the six ACGME dimensions, the seven models exhibit a similar relative ordering: patient care (PC) and professionalism (PROF) are the strongest, medical knowledge (MK), systems-based practice (SBP) and interpersonal and communication skills (ICS) occupy the middle range, and practice-based learning and improvement (PBLI) is the lowest in every model. PBLI completion does not exceed 30\% in any model evaluated, and falls below 20\% in both medical-domain models. The relative ordering across the six dimensions is therefore stable across model families, with PBLI emerging as a shared weak spot. A plausible explanation is that PBLI items typically require unprompted self-reflection, error acknowledgment or recognition of one's own knowledge limits, behaviours that are less likely to be elicited by current instruction-tuning objectives than the more directly cued actions assessed under PC or PROF.

\subsection{Effect of Context Length on Performance}
\label{sec:context_length}

In each evaluation, the assessed model's input context grows as the simulation run unfolds. This raises a natural question of whether the model will suffer from the `lost in the middle' problem~\cite{liu2024lost}, where, during continuous interactions with the environment, the extended context might become a burden or be forgotten by the model rather than a effective benefit for information gathering.

Figure~\ref{fig:results}c shows the overall relationship between the rubric completion rate and the assessed model's input context length across all models, and Figure~\ref{fig:results}d shows each model's detailed performance. In each panel, faint background points denote individual runs, and black dots denote binned summaries. These summaries are generated by sorting runs by input length and dividing them into ten equally sized bins, each containing 10\%, plotted at their respective mean input length and mean rubric completion rate. A trend line connects these binned means, accompanied by a shaded band showing the 95\% confidence intervals.

Contrary to the `lost in the middle' hypothesis, the completion rate increases monotonically with input length for most models with ample context windows, and this trend holds consistently across different model families. We observe no performance degradation at the upper end of the context length. This is expected, as the longest cases in our benchmark reach only about 40,000 tokens, which is far below the  maximum context windows length of current models (1M for the four general-purpose models, 256K for Qwen-3.5, and 128K for MedGemma; Table~\ref{tab:overall_performance}). This reflects that the effective context windows of modern mainstream models are fully adequate for handling most interactive medical SP scenarios. A longer context naturally reflects a more thorough clinical exploration and careful evidence-gathering process. On the same case, a model accumulates a longer context precisely because it carries out more rounds of history-taking, examinations, and test-ordering, thereby surfacing more of the information the case affords, which is exactly what the rubric credits.

Baichuan-M3 serves as a notable exception. Compared to other models, its official maximum context window is 40,960 tokens, closely matching the longest context generated in our simulations. As expected, its completion rate does decline at the upper end of the range (Figure\ref{fig:results}d), which is consistent with the longest scenarios approaching or exceeding its capacity. This suggests that current medical-specific LLMs, often optimized primarily for static QA tasks that typically involve short contexts, tend to overfit to short-context scenarios. Inadvertently, this narrow optimization may hurt  their long-context understanding capabilities, rendering them inadequate for complex, multi-step clinical decision-making.

\subsection{Evaluation Results across Specialties}
\label{sec:specialty_performance}

\begin{figure}[tbp]
    \centering
    \includegraphics[width=1\linewidth]{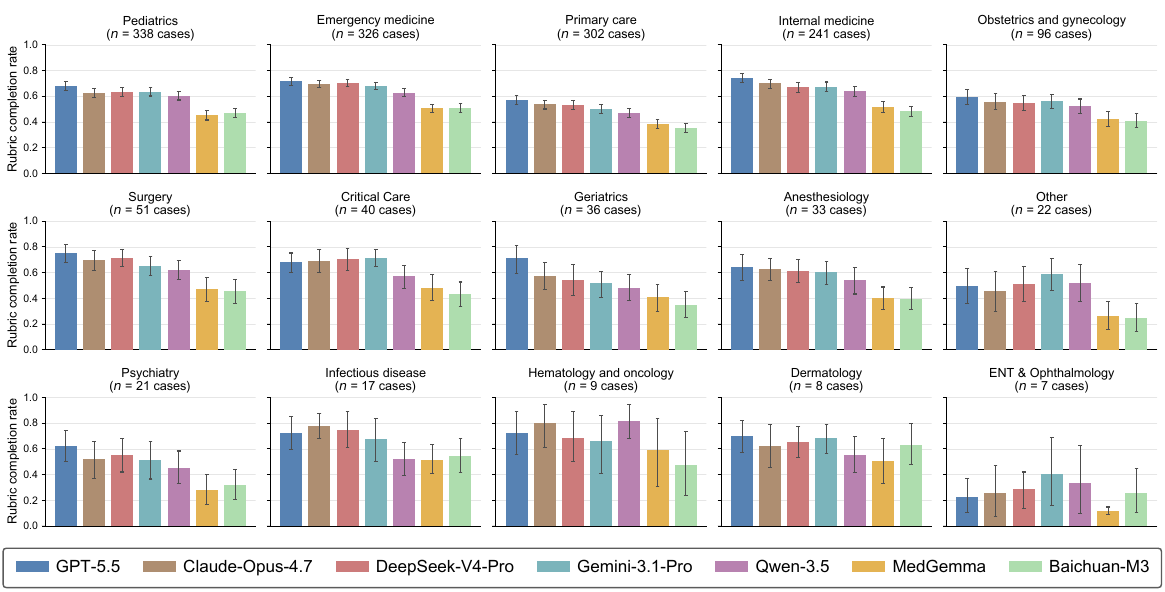}
    \caption{\textbf{Per-specialty performance on \benchname.} Bars are the per-case macro rubric completion rate of each model, and error bars are 95\% CIs. The number of cases per specialty is indicated in the panel header. Results are shown for the 15 specialties with sufficient case volume; the remaining two specialties of the 17-specialty taxonomy contained too few cases for stable per-model estimates and are omitted.}
    \label{fig:results_specialty}
\end{figure}

To examine whether model performance is uniform across various clinical contexts, we analyze the specialty-level rubric completion rate in \benchname. 
Although our cases span 17 different specialties, several specialties contain too few cases for stable per-model estimates; we therefore restrict the specialty analysis to the 15 specialties with at least five cases.
Figure~\ref{fig:results_specialty} shows the completion rate of each model per specialty.
 
Across all specialties, the ranking of model families largely follows the overall results in (Section~\ref{sec:overall_performance}). GPT-5.5 attains the highest score in 10 of the 15 specialties, with the lead passing to another general-purpose model in the remainder, most notably in Critical Care, where Gemini-3.1-Pro reaches 71.4\%, narrowly ahead of DeepSeek-V4-Pro (70.5\%) and GPT-5.5 (68.0\%). 
The two medical-tuned models take the lowest two positions in 12 of the 15 specialties.
In these 12 specialties, the general–medical gap~(the weakest general-purpose model minus the best medical-tuned model) ranges from 6.9 points in Hematology and oncology to 19.5 points in the Other category.

The three exceptions (Infectious disease, Dermatology and ENT \& Ophthalmology) are the three smallest specialties ($n\leq17$), where Baichuan-M3 rises into the range of the general models and the wide confidence intervals make the within-specialty ordering unreliable.

Absolute performance varies markedly across specialties. Among the higher-volume specialties ($n\geq36$), mean completion averaged across all seven models is highest in Emergency medicine (63.4\%), Internal medicine (63.1\%), Surgery (62.4\%) and Critical Care (61.0\%), and lowest in Primary care (47.8\%), Geriatrics (51.2\%) and Obstetrics and gynecology (51.6\%). The two extremes of the full distribution fall in the smallest specialties: ENT \& Ophthalmology has the lowest mean (26.8\%) and Hematology and oncology the highest (67.9\%), though both rest on fewer than 20 cases with correspondingly wide intervals. This ordering is largely preserved across individual models. One plausible explanation is that acute, protocol-driven specialties like Emergency medicine and Surgery provide clearer cues for action selection, whereas specialties marked by multi-system, longitudinal or psychosocial complexity like Primary care and Geriatrics require the model to integrate context across a wider range of considerations within a single encounter.

\subsection{Comparative Analysis of General-Purpose and Medical-Specialized Models}
\label{sec:medical_vs_general}

Across different view of the results, the two medical-specialized models, Baichuan-M3 and MedGemma, underperform all five general-purpose systems. On the micro metric they reach only 40.0\% and 39.5\%, at least 11 points below the weakest general model, Qwen-3.5 (51.5\%), and more than 20 points below GPT-5.5 (60.4\%; Table~\ref{tab:overall_performance}). The deficit is not confined to particular contexts: the two medical models occupy the lowest two positions in 12 of the 15 specialties (Figure~\ref{fig:results_specialty}) and trail on all six ACGME competencies rather than on any single one. The per-case score distribution shows the same pattern from a different angle (Figure~\ref{fig:results}a): the general models, and GPT-5.5 in particular, place a large fraction of cases at high completion and have distributions shifted toward the ceiling, whereas the medical models concentrate in the mid-to-low range and produce comparatively few high-scoring runs.The two medical models also carry the smallest context windows in our panel, 41K tokens for Baichuan-M3 and 128K for MedGemma, against 256K to 1M for the general models (Table~\ref{tab:overall_performance}); on the longest cases Baichuan-M3 therefore operates at the edge of its context capacity, which compounds its disadvantage.

Taken together, these results suggest that current mainstream medical benchmarks heavily favor static, single-turn question answering, causing most domain-specialized medical models to overfit to these formats. Consequently, this compromises their ability to navigate the more realistic, multi-turn, partially observed, and sequential decision-making inherent in actual clinical encounters. In contrast, general-purpose models, equipped with broad medical knowledge, robust long-horizon agentic capabilities, and superior instruction-following skills, prove significantly better suited for the interactive scenarios evaluated by \benchname. This underscores the pivotal role of \benchname in steering the future development of clinical LLMs, shifting the paradigm from static knowledge retrieval to more dynamic, long-context clinical reasoning.

\subsection{Analysis of the Effectiveness of Test-Time Scaling Strategies}

The main results reported above were obtained without complex inference scaling~\cite{wu2024inference}, in which the model produces its answer directly at each turn. A natural question is whether applying latest test-time compute scaling strategies to the strongest model, GPT-5.5, can further improve its performance. Given the substantial computational costs associated with these strategies, we conducted this experiment on \benchname-Verified, the high-quality, human-validated subset described in Section~\ref{sec:human_validation}, to ensure reliable conclusions within a manageable budget. 

We examine two representative test-time scaling strategies: (i) \emph{Best-of-$N$} sampling (here $N=5$), which runs the same case $N$ times independently under identical settings and aggregates the key decisions across the $N$ trajectories by self-consistency~\cite{self-consistency} (majority vote) to form the final answer; and (ii) multidisciplinary team (MDT) consultation~\cite{medagents}, a multi-agent pipeline that organizes GPT-5.5 into five specialist-clinicians roles deliberating in parallel, whose assessments are then consolidated by a single integrating role.

\begin{figure}[!t]
    \centering
    \includegraphics[width=\linewidth]{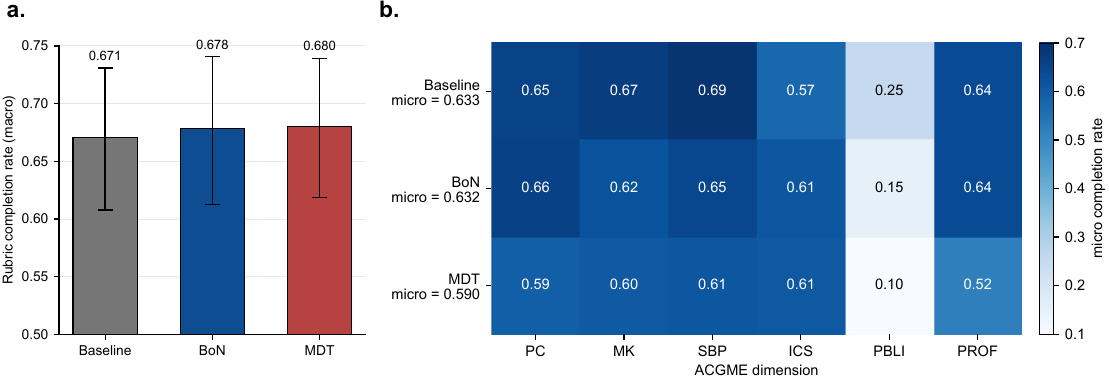}
    \caption{\textbf{Effect of test-time compute strategies on GPT-5.5, evaluated on the human-annotated subset.} \textbf{a.} Macro rubric-completion rate for the baseline, Best-of-$N$ (BoN, $N=5$) and MDT. \textbf{b.} Per-dimension micro completion rate across the six ACGME competencies (PC, MK, SBP, ICS, PBLI, PROF) for the three strategies.}
    \label{fig:test_time_scaling}
\end{figure}

As shown in Figure~\ref{fig:test_time_scaling}a, the single-pass baseline reaches a macro completion rate of 67.1\%, Best-of-5 with self-consistency aggregation reaches 67.8\%, and the five-specialist MDT reaches 68.0\%. 
Although MDT attains the nominally highest score, the differences among the three strategies are less than one percentage point and lack statistical significance. This suggests that neither test-time scaling through additional sampling nor multi-agent multidisciplinary collaboration produces a discernible overall improvement on this interactive clinical benchmark, implying that introducing more complex agentic prompting strategies cannot overcome the fundamental constraints of the base LLMs' intrinsic capabilities.

When we decompose performance into the six ACGME competencies by micro completion rate (Figure~\ref{fig:test_time_scaling}b), the only dimension in which both strategies consistently improve over single-pass is interpersonal and communication skills (ICS, 0.57$\to$0.61), in line with the intuition that additional deliberation and multi-role discussion benefit communication; on the remaining dimensions, however, both strategies are largely flat or lower. 
We attribute this to multi-agent prompting strategies instead introducing overconfidence, which leads the model to commit to decisions prematurely on some dimensions and thereby miss key information that would otherwise have been revealed in later turns (see Supplementary Figure~\ref{fig:supp_case_study_tts} for a representative case study).

\subsection{Case Study}

To make these patterns concrete, we further demonstrate three representative SP encounters, providing full turn-by-turn trajectories with each rubric item marked as completed or not.

\noindent\textbf{The clinician agent satisfies the majority of rubric items, yet small omissions remain.}
In an acute ischaemic stroke case (Supplementary Figure~\ref{fig:supp_case_study_stroke}), GPT-5.5 runs the entire initial-assessment phase within the clinically mandated window, establishing last-known-well, obtaining a finger-stick glucose, ordering appropriate labs and imaging, performing a focused NIHSS, and reaching a correct thrombolysis decision, scoring at or near ceiling on patient care, medical knowledge and systems-based practice. The two items it misses are fine-grained: it gives a 20\,mg rather than the guideline-mandated 10\,mg initial dose of labetalol, and it does not document explicit risk--benefit consent before alteplase. Both occur \emph{within} actions the model otherwise performs correctly, illustrating that even a near-ideal trajectory loses points at the finer level of protocol execution that single-turn benchmarks never probe.

\noindent\textbf{Adequate information gathering does not guarantee optimal care recommendations.} A prenatal nutrition counselling case (Supplementary Figure~\ref{fig:supp_case_study_nutrition}) exposes a gap between data collection and management. GPT-5.5 gathers a detailed quantitative dietary history, converting an open-ended recall into per-source intake estimates, yet scores only 3/5 on patient care and 2/7 on interpersonal and communication skills: it never states the recommended fish-intake guideline, does not address how preparation affects contaminant exposure, and leaves the patient's two closing actionable questions unanswered. The model collects enough information to give concrete guidance but delivers only vague cautions, a disconnect between gathering and acting that is invisible to any evaluation scoring a final answer alone.

\noindent\textbf{Multi-agent deliberation can introduce early-exit failures.}  Under the MedAgents scaffold (Supplementary Figure~\ref{fig:supp_case_study_tts}), in which five GPT-5.5 subspecialists emit proposals aggregated by majority vote, a PICU case of a 2-year-old with altered mental status terminates prematurely. At the seventh turn, three subspecialists vote to end the encounter on the premise that the child is already stabilised, overriding the emergency-medicine and critical-care agents who note that core resuscitation has not yet occurred; the 3-to-2 majority prevails and the run closes. A cluster of basic patient-care items the dissenters explicitly demanded, including a fluid bolus, bedside glucose, venous blood gas, and naloxone, therefore go unreached. Convening multiple perspectives makes the system more inclined to judge the situation already handled, showing that the additional inference afforded by multi-agent deliberation is not always beneficial and that the aggregation mechanism itself can create failures, consistent with the flat test-time scaling results above.

\section{Discussion}
In this study, we present \benchname, an interactive benchmark that evaluates large language models as clinical agents under dynamic educational clinical environments derived from standardized patient (SP) cases. By converting peer-reviewed SP teaching materials into executable, closed-loop scenarios, \benchname assesses an agent's ability to actively elicit information, track an evolving patient state, and commit to time-sensitive actions, while being objectively and quantifiably scored at each step against expert-defined reference actions. The key contributions of this study are as follows:

\textbf{The first sequential clinical decision-making evaluation dataset built from educational SP cases.} To our knowledge, \benchname is the first benchmark to systematically mine the rich repository of SP cases in medical education for LLM evaluation. By repurposing traditional medical training materials, we create a novel resource that evaluates LLMs as sequential clinical decision-makers in dynamic clinical scenarios with quantifiable scoring. Specifically, we curate 1,638 cases across 17 clinical specialties, carrying 24,602 expert-defined rubric items mapped to the six ACGME competencies. This curation is driven by a fully automated pipeline, whose reliability has been validated through clinician review of a 100-case subset. Compared to recent similar interactive benchmarks~\cite{johri2025evaluation, agentclinic,tu2025towards,agenthospital, qiu2025evolving} that predominantly focus on diagnostic history-taking or online consultations, \benchname leverages the inherent advantages of SP cases to offer greater scenario diversity, measure a broader spectrum of clinical competencies, and ensure peer-reviewed high quality.

\textbf{A dynamic, multi-agent evaluation framework.} We design a closed-loop interactive system in which the model under evaluation engages a patient agent and an environment controller across multiple time steps under a standardized state-transition protocol. This framework shifts evaluation from outcome-only judgement to process-level assessment, scoring whether a model gathers the right information at the right moment, updates its judgement as the patient state changes, and executes appropriate timely interventional actions, organized along six ACGME clinical capability dimensions.

\textbf{A systematic evaluation of a broad range of LLMs.} We evaluate a representative set of state-of-the-art systems on \benchname, spanning frontier closed-source models, leading open-source reasoning models, and medical-domain models, providing the first process-level comparison of these systems under a unified interactive clinical protocol and a characterization of their recurrent failure modes.

We summarize the main findings on \benchname as follows:

\textbf{Strong medical knowledge does not translate into reliable real-world interactive clinical practice.} Our findings reveal a substantial gap between the medical knowledge exhibited by current LLMs and their ability to apply it in interactive clinical practice. Although recent LLMs perform strongly on static medical question-answering benchmarks~\cite{singhal2023large, singhal2025toward, zuo2025medxpertqa}, this does not imply readiness for real-world clinical use. In \benchname, which uses standardized patient cases from medical education, the best-performing model, GPT-5.5, achieved only a 60.4\% rubric completion rate. Thus, failures to complete necessary actions even in structured educational scenarios raise serious concerns about LLM reliability in more complex real-world settings. These concerns extend beyond knowledge or diagnostic accuracy, as safe care also requires interactive actions such as asking key history questions, ordering indicated tests, communicating return precautions, and escalating care when needed. These process-level failures are difficult to detect in static QAs, where models are evaluated primarily on final answers rather than care delivery. As a result, strong performance on knowledge-based benchmarks may inflate perceived clinical readiness and encourage inappropriate overreliance. Current LLMs should therefore remain under strict human supervision and be deployed only as assistive tools in bounded clinical workflows, rather than as autonomous interactive clinician agents.

\textbf{Medical-domain models do not outperform general-purpose LLMs.} 
The two medical-specialized models, Baichuan-M3 and MedGemma, rank at the bottom of every comparison, with micro completion rates of 40.0\% (Baichuan-M3) and 39.5\% (MedGemma); even the stronger of the two falls 11.5 points below the weakest general-purpose model (Qwen-3.5, 51.5\%) and 20.4 points below GPT-5.5 (60.4\%). This shortfall is distributed evenly across all six ACGME competencies rather than driven by any single dimension, indicating a uniform offset in interactive competence.  A plausible reason is that mainstream medical benchmarks heavily favor static, single-turn question answering, causing domain-specialized models to overfit to these formats. Consequently, their ability to navigate the multi-turn, sequential decision-making inherent in realistic clinical scenarios is severely compromised. Our \benchname is specifically designed to highlight this discrepancy and steer the future development of clinical LLMs, shifting the paradigm from static knowledge retrieval to dynamic, long-context clinical reasoning.

\textbf{Model rankings are stable across specialties, while absolute scoring is not.} The ordering of model families is largely preserved across the 15 specialties: GPT-5.5 ranks first in 10 of them, the general-purpose models hold the top positions throughout, and the two medical-tuned models remain at the bottom in 12 of the 15, with a general-to-medical gap that stays positive and ranges from 7.7 points in Geriatrics to 15.5 points in Surgery. Absolute performance, by contrast, varies substantially: acute, protocol-driven settings such as Emergency medicine, Surgery, Internal medicine and Critical Care are the easiest, whereas specialties demanding multi-system, longitudinal or psychosocial integration, such as Primary care, Obstetrics and gynecology and Geriatrics, are the hardest, exposing a systematic failure gap shared across current LLMs.

\textbf{Additional test-time scaling does not improve interactive performance.} Allocating additional test-time compute to the strongest model, GPT-5.5, yields no discernible overall gain: the single-pass baseline scores 67.1\%, Best-of-5 with self-consistency aggregation 67.8\%, and a five-specialist MedAgents consultation 68.0\%, differing by at most 0.9 points and statistically indistinguishable. The only competency that consistently benefits is interpersonal and communication skills (ICS, from 0.57 to 0.61), while the remaining dimensions are flat or slightly lower. We attribute this to the overconfidence introduced by multi-agent deliberation, which leads the model to prematurely stop interaction, thus, missing information that later turns would otherwise have surfaced.

\textbf{Limitations:}
\benchname is built on standardized-patient and simulation materials authored for medical education, and evaluation is conducted through LLM-driven patient and environment agents. While this design provides expert-defined reference actions and reproducible trajectories, the resulting scenarios remain idealized teaching simulations rather than real clinical encounters, and a gap therefore exists between performance on \benchname and readiness for live clinical deployment. \benchname is best understood as a standardized, pre-deployment assessment of interactive clinical competence, analogous to objective structured clinical examinations in medical training, rather than as direct evidence of bedside readiness.
\section{Methods}

\begin{figure}[h]
    \centering
    \includegraphics[width=1\linewidth]{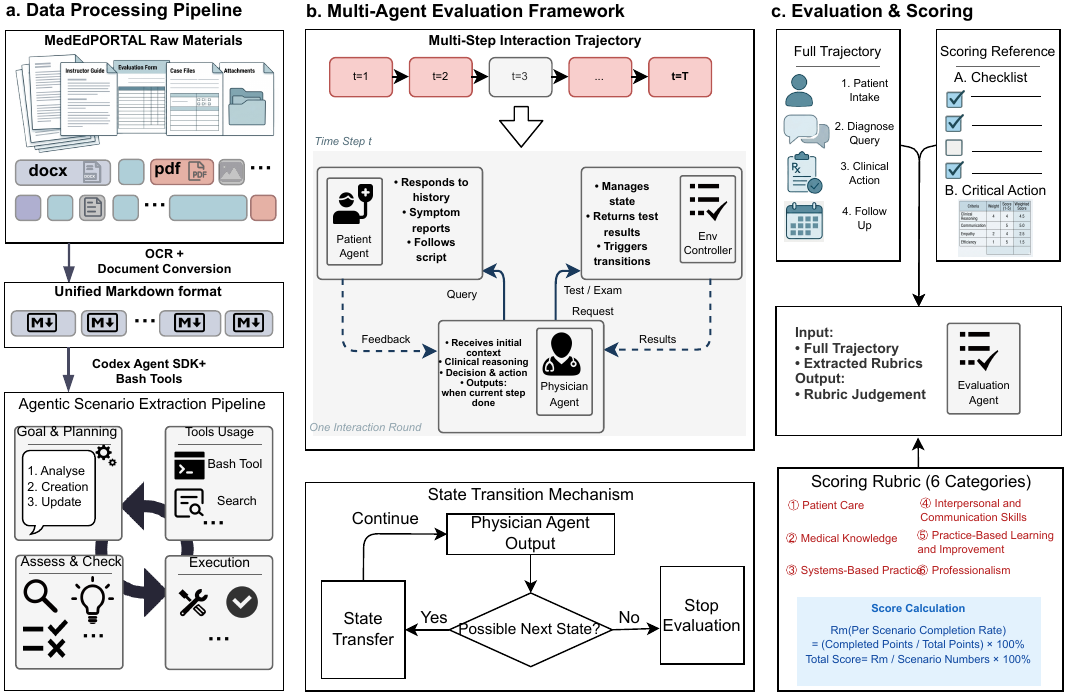}
    \caption{Overview of the {\benchname} data processing and evaluation pipeline. \textbf{a.} shows the data processing pipeline, which transforms heterogeneous MedEdPORTAL teaching materials into role-specific scenario materials via OCR-based unification and an agentic extraction pipeline. \textbf{b.} shows the multi-agent evaluation framework, in which a clinician agent interacts with a patient agent and an environment controller across multiple clinical states under a standardized state-transition protocol. \textbf{c.} shows the evaluation and scoring procedure, where an evaluator agent compares the clinician agent's full trajectory against rubric-based scoring references derived from the original materials, and produces a scoring-point completion rate across the six ACGME core competencies. }
    \label{fig:methods_overview}
\end{figure}

This section describes the development of \benchname, including the data curation pipeline, the multi-agent evaluation framework, and the implementation of evaluation metrics. All text prompts used are provided in the supplementary.

\subsection{Data Curation}
\label{sec:data_curation}

\begin{figure}[h]
    \centering
    \includegraphics[width=0.8\linewidth]{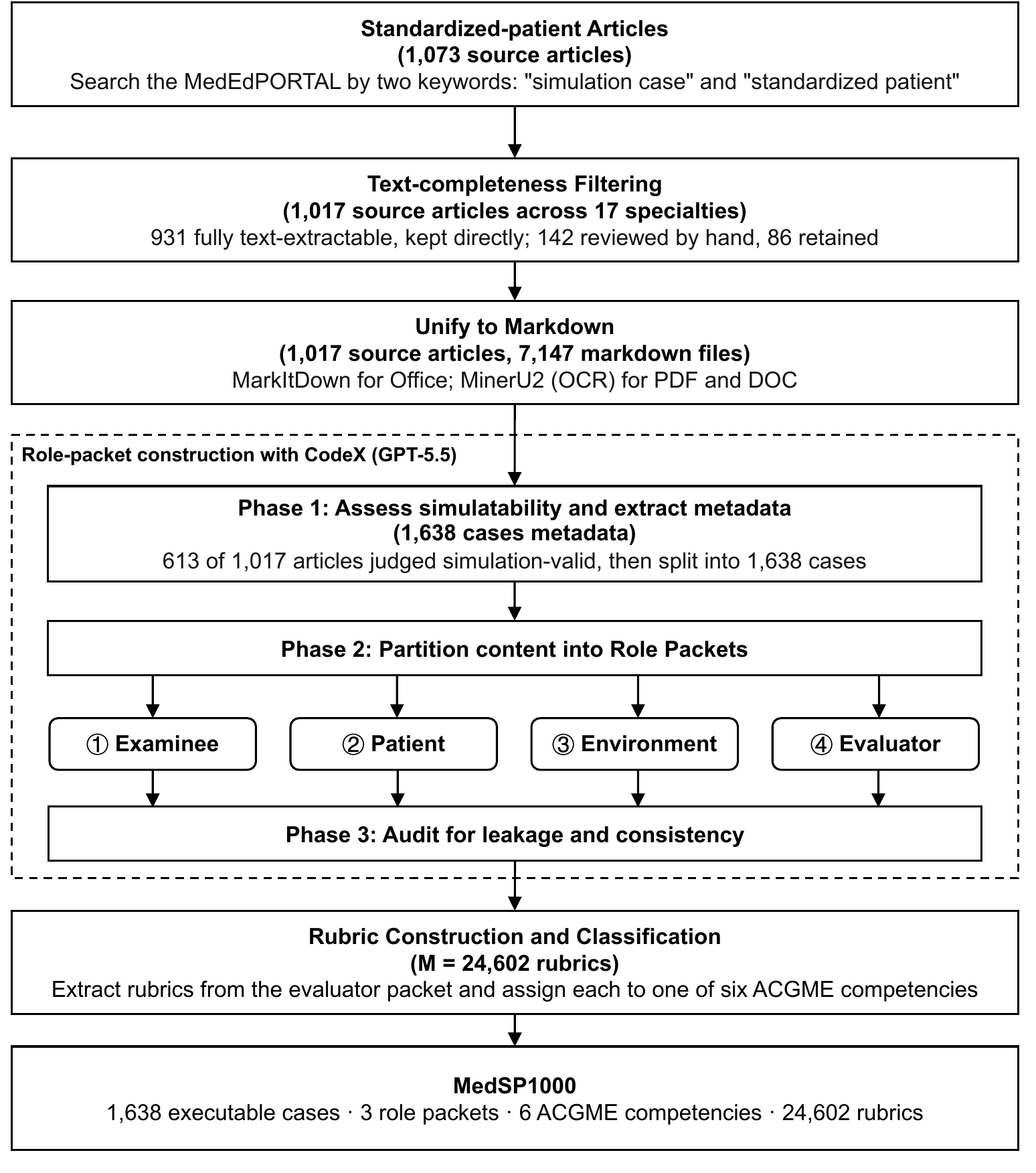}
    \caption{\textbf{Overview of the MedSP1000 data curation pipeline.} MedEdPORTAL source articles are filtered, unified into Markdown, and processed by a three-phase Codex pipeline that partitions each case into role packets, with the evaluator packet converted into a scoring rubric.}
    \label{fig:methods_flow_chart}
\end{figure}

As illustrated in Figure~\ref{fig:methods_overview}a, the source data for \benchname are collected from MedEdPORTAL\footnote{\url{https://www.mededportal.org/}}, an open-access repository of teaching and assessment resources for professional medical education. As shown in Figure~\ref{fig:methods_flow_chart}, to construct the source articles into a sequential decision-making benchmark, we implement the following pipeline:


\textbf{Source article collection and preprocessing.} We first retrieved SP case articles from MedEdPORTAL by searching the keywords 'simulation case' and 'standardized patient,' then downloaded the main texts and supplementary materials. This yields a total of unique \textbf{1,073} source articles, accompanied by \textbf{22,244} supplementary files. 
Among these files, text-only files account for \textbf{69.3\%} (\textbf{15,423} files), while the remainder are multi-modal materials including interactive courseware, images, audio, and video. A detailed statistic by modality and file type is provided in Supplementary Figures~\ref{fig:supp_format_distribution_modality} and~\ref{fig:supp_format_distribution_extension}.

As our pipeline is text-based, a source article is only usable if all critical information is provided as text. Therefore, we then screen the \textbf{1,073} multi-modal sources in two steps. First, we directly retain \textbf{931} articles whose attachments are exclusively in text-extractable formats (PDF, DOCX, DOC, PPT, PPTX, HTML, TXT, and Markdown). Second, the remaining \textbf{142} articles, which contain at least one non-text-processable attachment, undergo manual review to determine if these files hold indispensable simulation data. Articles are retained if the non-text attachments are non-essential~(\textbf{86} articles) and discarded otherwise~(\textbf{56} articles). This process further yields \textbf{1,017} articles for inclusion.

Finally, we convert each retained article into a uniform Markdown format that can be read directly. Documents with explicit structural markup, like DOCX, PPT, PPTX, are converted into MarkItDown~\cite{markitdown} to preserve heading hierarchy, lists and tables; DOC and PDF files are processed with MinerU2~\cite{mineru2} for layout restoration and OCR-based text extraction; and plain-text formats~(TXT, Markdown) are passed through unchanged. This stage produces \textbf{7,147} Markdown files, grouped by source article.

\textbf{Constructing role-specific packets.} In this phase, we partition each source article's Markdown text into four role-specific packets: a scenario initialization, a patient script, environment-controller materials, and a scoring rubric. These correspond to the four participants in the following multi-agent evaluation framework: the \emph{assessed model}, the \emph{patient agent}, the \emph{environment controller}, and the \emph{evaluator}. Each packet defines the exact information its corresponding participant can access during the evaluation, for instance, the scoring rubric serves as input solely for the evaluator.

This partitioning must satisfy two key requirements. The first is the strict prevention of \emph{information leakage}: no agent should access privileged information assigned to another role which could compromise the integrity of the evaluation. For example, the assessed model is restricted to the information available to a clinician at the onset of the encounter, can progressively access more content dynamically revealed during the interaction, and must never non-causally access future disease progression, untriggered test results, or the evaluator's reference answers. The second is \emph{fidelity to the source}: the original SP materials must remain entirely unaltered, free from any additions, deletions, or revisions. In particular, no test results, disease progression, or scoring criteria absent from the original text may be introduced.

To meet both requirements, we use Codex, an off-the-shelf agentic system powered by GPT-5.5, to perform the construction in three phases. Codex is equipped with tool-calling capabilities, allowing it to execute bash commands, manage the file system, and invoke Python for structured parsing and cross-file consistency checks:
\begin{itemize}
\setlength\itemsep{3pt}

    \item \textbf{Assessing simulatability and extracting metadata} Using the prompt shown in Supplementary Section~\ref{suppsec:phase1}, we first prompt Codex to read all Markdown files of a source article and record the content and intended use of each file. It then judges whether the article can support a text-based clinical simulation and extracts scenario metadata such as specialty, urgency, and whether the article should be split into multiple independent encounters. Following HealthBench Professional~\cite{healthbench-pro}, we classify each case into one of 17 clinical specialties. Of the \textbf{1,017} articles entering this pipeline, \textbf{613} are judged simulation-valid. Because a single source article may contain several distinct standardized-patient scenarios, we then split each article's metadata so that every metadata entry corresponds to exactly one scenario, yielding a total of \textbf{1,638} executable cases.
    

    \item \textbf{Partitioning content into role-specific compacts}  As shown in the prompt in Supplementary Section~\ref{suppsec:phase2}, for each remaining article we prompt Codex to perform a copy-and-delete operation: it creates the four directories (\texttt{scenario\_initialization}, \texttt{standardized\_patient\_actor}, \texttt{environment\_\allowbreak controller}, and \texttt{evaluator}), copies the relevant source Markdown into each in full, and then deletes the passages that each role must not see according to their visibility rules. By doing this, every retained sentence remains directly traceable to the original paragraphs rather than being a rewritten paraphrase. This bypasses the hallucination introduced by text generation and allows for easy verification against the source text. A case of this operation can be found in Supplementary Figure~\ref{fig:supp_case_copydelete}.
    

    \item \textbf{Auditing for leakage and consistency}  As shown in the prompt in Supplementary Section~\ref{suppsec:phase3}, we prompt Codex to self-reflect and audit its outputs from the preceding phase. It checks whether the role-specific packets retain any information semantically equivalent to an evaluator scoring item, whether the patient packet is sufficient and accurate for portrayal, and whether the splitting into encounters is internally consistent. Based on the audit, it iteratively corrects any identified anomalies.
\end{itemize}

\noindent\textbf{Rubric Parsing and Categorization.} The rubric packet contains the original educators' scoring points. Although they are itemized, their inconsistent free-text formatting makes uniform automated processing difficult. Thus, we extract the raw text into a standardized list of scoring items.  As shown in the prompt in Supplementary Section~\ref{suppsec:rubric_extraction}, we prompt Codex to read a case's evaluator packet, identify every passage that expresses a scoring point, and output it as a discrete item. To preserve source granularity, we instruct the model to copy the wording verbatim rather than rewriting or merging it. Each item is then mapped to one of the six ACGME core competencies~\cite{edgar2018milestones,edgar2020milestones}. This yields a competency-organized rubric for each case. The rubric is constructed once and applied unchanged to every model evaluated on that case, ensuring all models are scored against an identical set of items for direct comparability. This phase results in a total of 24{,}602 scoring criteria.

\subsection{Evaluation Framework}
\label{sec:evaluation_framework}

In this section, we introduce the implementation details of our evaluation framework. \benchname converts each SP case from a static teaching document into an interactive simulation where clinical LLMs can be evaluated under a standardized protocol (see Supplementary Section~\ref{sec:case_study} for representative simulation instances).

The framework is built around two simulation agents that jointly reconstruct the clinical environment: a \emph{patient agent} that handles communicative interaction, and an \emph{environment controller agent} that resolves the model's actions and governs how the scenario progresses over time. Afterwards, an evaluation agent is introduced to judge the trajectory based on rubrics. The model under evaluation interacts with the environment through a structured interface.

\noindent\textbf{Patient agent.} The patient agent plays the standardized patient, engaging the model under evaluation in multi-turn dialogue throughout history-taking, symptom inquiry, and interpersonal communication. At each turn it receives the model's output action and produces an in-character response dynamically. The patient agent is conditioned on the processed patient packet and instructed to strictly constrain its output, disclosing only information consistent with the current stage, following the predefined scripts. It is driven by DeepSeek-V4-Pro under the system prompt given in full in Supplementary Section~\ref{suppsec:prompt_sp}.

\noindent\textbf{Environment controller agent.} The environment controller represents the available interactive clinical environment beyond the patient, including other clinical staff such as nurses and consulting specialists, monitoring devices, laboratory and imaging services, and the results of examinations, tests, and treatments. It carries two responsibilities:
(\textit{i}) It responds to the model's actions, returning the corresponding results, like physical examinations, test orders, procedures, treatments, and dispositions, when the source materials specify them for the current stage; actions for which the source materials provide no result are marked as unsupported rather than fabricated. (\textit{ii}) It manages temporal progression. Each scenario is organized as a fixed sequence of discrete \emph{clinical states} defined by the source materials, with boundaries set by points at which the patient's critical longitudinal condition changes, \emph{e.g.}, an initial presentation, a deterioration, or a new state following an intervention). When the model signals that it considers the current state complete, the controller checks the source materials to determine whether a subsequent state exists and whether to transition to it. If so, it releases the initial information for the next state, otherwise, it ends the run. The assessed model cannot advance the state on its own, and the framework imposes no fixed maximum number of turns. Termination is determined solely by whether the clinical states provided in the environment controller packet have been exhausted. The controller is likewise driven by DeepSeek-V4-Pro, and its full system prompt is provided in Supplementary Section~\ref{suppsec:prompt_env_controller}.

\noindent\textbf{Evaluator agent.} The evaluator scores the completed dynamic decision-making processes. Upon termination of a run, it receives the full interaction trajectory alongside the rubric packet. For each rubric item, it produces a binary judgment indicating whether the model successfully satisfied that criterion during the session. By utilizing a shared, objective rubric, the evaluator guarantees uniform assessment across all models. Its full system prompt is provided in Supplementary Section~\ref{suppsec:prompt_evaluator}, and the metrics computed from its judgements are defined in Section~\ref{sec:method_eval_metrics}.

\noindent\textbf{Interaction interface.} At the start of a SP simulation run, the assessed LLM receives only the content in the scenario initialization packet, describing the role and responsibilities it is playing, the clinical setting, and the patient's presenting complaint and objective situation. At each turn, the model is instructed to produce a structured output with three fields: \texttt{speak}, the utterance addressed to those present; \texttt{actions}, a list of clinical actions; and \texttt{eos}, a boolean signal indicating whether the model considers the current clinical state complete. The framework routes \texttt{speak} to the patient agent and \texttt{actions} to the environment controller, and exposes their responses back to the model on the next turn. Once all clinical states are exhausted, the full interaction trajectory is passed to the evaluator agent. The full interaction interface prompt is provided in Supplementary Section~\ref{suppsec:prompt_examinee}.


\subsection{Evaluation Metrics}
\label{sec:method_eval_metrics}


Denoting a given model $m$ and case $s$, let $C_{m,s}$ be the number of rubric items judged completed and $T_s$ the total number of items in the rubric of $s$, we adopt the following metrics to measure models' performance.

\noindent\textbf{Rubric completion rate.} The case is the primary unit of aggregation.
We define the per-case \emph{rubric completion rate} as

\begin{equation}
r_{m,s} = \frac{C_{m,s}}{T_s},
\label{eq:per_scenario_rate}
\end{equation}
that is, the fraction of rubric items completed in that run. A model's overall score is the macro average of its per-case completion rates across all cases it has run,
\begin{equation}
R_m = \frac{1}{|S|}\sum_{s \in S} r_{m,s},
\label{eq:overall_score}
\end{equation}
where $S$ is the set of evaluated cases. Every case contributes equally to $R_m$ regardless of how many rubric items it contains, so $R_m$ reflects average case-level performance rather than item-level performance.

\noindent\textbf{Competency- and specialty-level aggregation.} The same construction extends to subgroup analyses. For an ACGME competency $d \in \{\mathrm{PC},\mathrm{MK},\mathrm{SBP},\mathrm{ICS},\mathrm{PBLI},\mathrm{PROF}\}$, let $C^{(d)}_{m,s}$ and $T^{(d)}_s$ denote the completed and total rubric items of $s$ under dimension $d$. The per-case, per-dimension completion rate is $r^{(d)}_{m,s} = C^{(d)}_{m,s}/T^{(d)}_s$, and the model's competency-level score is the macro average
\begin{equation}
R^{(d)}_m = \frac{1}{|S_d|}\sum_{s \in S_d} r^{(d)}_{m,s},
\label{eq:competency_score}
\end{equation}
where $S_d = \{s \in S : T^{(d)}_s > 0\}$ restricts the average to cases that actually contain rubric items under dimension $d$.  Specialty-level scores are computed analogously, by restricting the macro average in equation~\eqref{eq:overall_score} to cases within a given specialty.

\subsection{LLM Baselines}
\label{sec:baselines}
Here, we introduce the baseline LLMs involved in our experiments:

\vspace{-6pt}
\begin{itemize}\setlength\itemsep{3pt}
    \item \textbf{Claude-Opus-4.7}~\cite{claude4-7}: Released by Anthropic in April 2026, Claude-Opus-4.7 is widely recognized for its strong capabilities on coding and agentic tasks, and is capable of handling complex, long-horizon workflows with minimal human supervision. We evaluated the model version \texttt{claude-opus-4-7} using the official API.
    
    \item \textbf{GPT-5.5}~\cite{gpt5-5}: GPT-5.5, released by OpenAI in April 2026, is the latest flagship of the GPT series. It is designed to plan, invoke tools, and verify its own outputs across multi-step real-world workflows, with notable strengths in agentic coding, computer use, and scientific research. We evaluated the model version \texttt{gpt-5.5} using the official API.
    
    \item \textbf{Gemini-3.1-Pro}~\cite{gemini-3-1-pro}: Developed by Google DeepMind and released in February 2026, Gemini-3.1-Pro is the latest iteration of the Gemini 3 series. As a natively multimodal model, it accepts a wide range of inputs including text, images, audio, video, and code, and is well-suited for tasks that involve cross-modal understanding. We evaluated the model version \texttt{gemini-3.1-pro-preview} using the official API.
    
    \item \textbf{DeepSeek-V4-Pro}~\cite{deepseekai2026deepseekv4}: DeepSeek-V4-Pro is a 1.6T-parameter Mixture-of-Experts (MoE) LLM developed by the DeepSeek AI and released in April 2026. Among current open-source models, it stands out for its top-tier performance on coding and agentic benchmarks, and substantially narrows the gap with leading closed-source counterparts. We evaluated the model version \texttt{deepseek-v4-pro} using the official API.
    
    \item \textbf{Qwen-3.5}~\cite{qwen3.5}: Qwen-3.5 is the latest open-source LLM series developed by the Qwen Team at Alibaba. We use its flagship open-weight variant, Qwen3.5-397B-A17B, a Mixture-of-Experts model with 397B total parameters and 17B activated per token. We deploy the model weights from Huggingface (\url{https://huggingface.co/Qwen/Qwen3.5-397B-A17B}), \texttt{Qwen/Qwen3.5-397B-A17B}, locally for evaluation.

    \item \textbf{MedGemma}~\cite{sellergren2025medgemma}: MedGemma is a variant of Gemma 3, which is optimized for the medical domain by Google DeepMind. Unlike the previously mentioned models, which are designed for general domains, MedGemma is a specialized medical LLM. In our evaluation, we use the 27B-parameter model weight from Huggingface (\url{https://huggingface.co/google/medgemma-27b-text-it}), \texttt{google/medgemma-27b-text-it}, locally for assessment.

    \item \textbf{Baichuan-M3}~\cite{baichuan-m3}: Baichuan-M3 is a 235B-parameter MoE medical-specific LLM developed by the Baichuan company, released in January 2026. Similar to MedGemma, it is a specialized medical LLM rather than a general-domain model. Unlike previous medical LLMs that focus on static question answering, Baichuan-M3 is explicitly trained to model the clinical decision-making process. We use the model weights from Huggingface (\url{https://huggingface.co/baichuan-inc/Baichuan-M3-235B}), \texttt{baichuan-inc/Baichuan-M3-235B}, deployed locally for evaluation.
\end{itemize}

We also evaluate GPT-5.5 under two representative test-time scaling strategies:

\begin{itemize}
    \item \textbf{Best-of-N.} Following the self-consistency decoding strategy~\cite{self-consistency}, we run each case $N$ times independently under identical settings (here $N=5$), sampling a diverse set of trajectories rather than committing to a single greedy decode. At each decision point, the key actions are aggregated across the $N$ trajectories by majority vote, and the most consistent choice is taken as the final answer. The intuition is that a difficult clinical decision typically admits multiple plausible reasoning paths that nonetheless converge on the same correct action, so marginalizing over sampled trajectories should suppress idiosyncratic errors.

    \item \textbf{MedAgents.} We adopt the multidisciplinary collaboration framework of \cite{medagents}, a training-free multi-agent pipeline that casts a single base model into multiple specialist-clinician roles engaged in a role-playing discussion. Concretely, GPT-5.5 instantiates five specialist agents that independently propose their own analyses for the current turn; a separate integrating role then summarizes these analyses and consolidates them into a single decision. This mirrors a real-world multidisciplinary team (MDT) consultation and is intended to elicit and combine the domain knowledge distributed across the base model. In our interactive setting, each specialist emits a \{speak, actions, eos\} proposal at every turn, and the integrating role aggregates them---including the decision of whether to end the encounter---by majority vote.
\end{itemize}

\subsection{Ethics declaration}
This study was constructed entirely from MedEdPORTAL, an open-access, peer-reviewed repository of health-professions teaching materials published under Creative Commons (CC BY, CC BY-NC or CC0) licences. The source materials are instructional standardized-patient and simulation cases and do not contain identifiable patient information. They were released by their original authors in compliance with the relevant ethical, legal, and regulatory requirements, and we accessed and used them solely under their original licence terms for non-commercial research, retaining the original licence and author attribution for each case. Our study involved no patient recruitment, no new collection of patient data, and no access to protected health information; accordingly, no informed consent process was required for the case data.

\section*{Data availability}
\benchname is derived from MedEdPORTAL, an open-access, peer-reviewed repository of health-professions teaching materials. The source standardized-patient and simulation articles are released by their original authors under Creative Commons licences\footnote{https://www.mededportal.org/author\#creativecommonslicensing} (CC BY, CC BY-NC, or CC0). We accessed and processed them solely under their original licence terms for non-commercial research, retaining the original author attribution for each case.
From these materials we constructed \benchname, comprising 1,638 executable scenarios with role-specific packets and frozen rubrics derived from 613 source articles (see Section~\ref{sec:data_curation}).
The constructed \benchname dataset is publicly available at \url{https://huggingface.co/datasets/byrLLCC/MedSP1000} under a CC BY-NC-SA licence, chosen to remain compatible with the non-commercial terms of the underlying sources. To allow every scenario to be traced back to its origin, a Supplementary CSV file lists all 613 source articles with their titles, original authors, and MedEdPORTAL URLs.

\section*{Code availability}
Source codes of this paper is released in \url{https://github.com/MAGIC-AI4Med/MedSP1000} with \textit{CC BY-SA} license.
{
\rev

\section*{Author Contributions}
All listed authors clearly meet the ICMJE 4 criteria. C.L. and P.Q. contribute equally to this work. C.W. and W.X. are the corresponding authors. Specifically, C.L., P.Q., Y.Z., Y.W., C.W., and W.X. all make contributions to the conception or design of the work, and C.L. and P.Q. further perform acquisition, analysis, or interpretation of data for the work. In writing, C.L. and P.Q. draft the work. Y.Z., Y.W., C.W., and W.X. review it critically for important intellectual content. All authors approve of the version to be published and agree to be accountable for all aspects of the work to ensure that questions related to the accuracy or integrity of any part of the work are appropriately investigated and resolved.

\section*{Acknowledgements}
This work is supported by the Scientific Research Innovation Capability Support Project for Young Faculty (ZYGXQNJSKYCXNLZCXM-I22), the National Natural Science Foundation of China (No. 24Z031503678), and the  Shanghai Municipal Special Program for Basic Research on General AI Foundation Models (Grant No. 2025SHZDZX026D12).

\section*{Competing interests}
The authors declare no competing interests.
}

\bibliographystyle{unsrt}
\bibliography{main} 
\clearpage
\section{Supplementary}

\renewcommand{\tablename}{Supplementary Table}
\renewcommand{\figurename}{Supplementary Figure}

\setcounter{figure}{0}
\setcounter{table}{0}

\DefineVerbatimEnvironment{PromptBlock}{Verbatim}{
    breaklines=true,
    breakanywhere=true,
    fontsize=\small,
    frame=single,
    framesep=3mm,
    rulecolor=\color{gray},
    commandchars=\\\{\}
}

\DefineVerbatimEnvironment{PromptBody}{Verbatim}{
    breaklines=true,
    breakanywhere=true,
    fontsize=\small,
    commandchars=\\\{\}
}

\newtcolorbox{promptbox}[2][]{
    enhanced,
    breakable,
    colback=gray!15,
    colbacktitle=gray!40,
    coltitle=black,
    fonttitle=\bfseries\small,
    title=#2,
    label=#1,
    boxrule=0pt,
    arc=1pt,
    left=2mm, right=2mm, top=1mm, bottom=1mm,
    attach boxed title to top left={xshift=0pt, yshift=0pt},
    boxed title style={boxrule=0pt, arc=0pt, sharp corners},
    title filled,
}

\begin{figure}[h]
    \centering
    \includegraphics[width=0.95\linewidth]{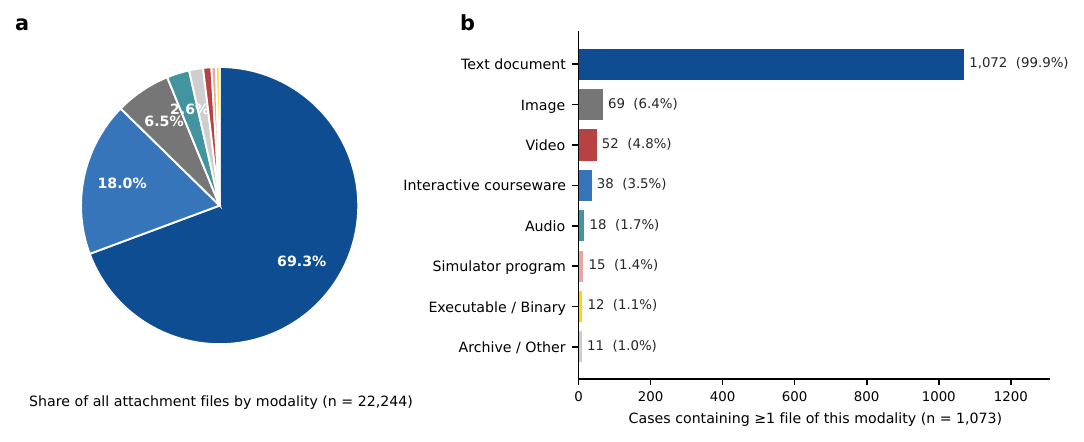}
    \caption{Coarse-grained modality-level distribution of attachments in the MedEdPORTAL source corpus. Attachments are grouped into broad categories, including text, image, audio, video and programmatic resources.}
    \label{fig:supp_format_distribution_modality}
\end{figure}

\begin{figure}[h]
    \centering
    \includegraphics[width=0.95\linewidth]{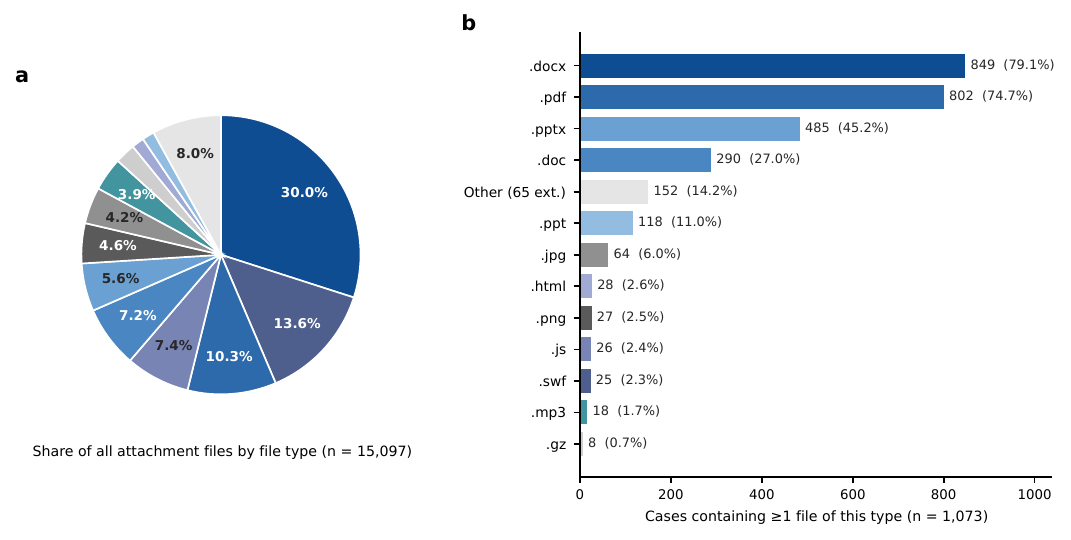}
    \caption{Fine-grained distribution of attachments by original file extension and source file type in the MedEdPORTAL source corpus.}
    \label{fig:supp_format_distribution_extension}
\end{figure}

\subsection{Full Results on the Human-Annotated Subset}
\label{suppsec:subset_results}

Supplementary Table~\ref{tab:subset_performance} reports the per-model rubric completion rate on the human-annotated subset of 100 cases, the high-quality subset constructed and verified by twelve clinicians that is also used for the test-time scaling experiment in the main text.

The results closely track those on the full benchmark (Table~\ref{tab:overall_performance}), supporting the reliability of the automated pipeline. First, the relative ordering of models is preserved: GPT-5.5 remains the strongest system on both the micro and macro metrics, the general-purpose models occupy the top positions, and the two medical-specialized models, Baichuan-M3 and MedGemma, remain at the bottom. Second, the general--medical gap persists at a similar magnitude, with GPT-5.5 (64.4\% micro) leading the best medical model (MedGemma, 49.0\% micro) by 15.4 points. Third, the per-competency profile is unchanged: practice-based learning and improvement (PBLI) is the weakest dimension for every model, not exceeding 16.7\% in any system, while patient care (PC) and professionalism (PROF) remain among the strongest. 

The principal difference is that absolute completion rates are higher on this subset than on the full benchmark across all models; this is expected, since these cases were filtered for construction quality and span a narrower, more cleanly-specified set of scenarios, but the consistent ordering and competency structure indicate that the full-benchmark conclusions are not driven by construction noise.

\begin{table}[!ht]
\renewcommand{\arraystretch}{1.3}
\footnotesize
\centering
\caption{Per-model rubric completion rate on the human-annotated subset of \textbf{100} scenarios. All values are percentages. PC, patient care; MK, medical knowledge; SBP, systems-based practice; ICS, interpersonal and communication skills; PBLI, practice-based learning and improvement; PROF, professionalism. For each competency, the micro completion rate is shown. \emph{Micro} pools all rubric items across the subset; \emph{Macro} is the per-scenario macro completion rate (mean over the 100 scenarios). \textbf{Bold} marks the best value in each column. In each cell, the top line is the point estimate and the bottom line is its 95\% confidence interval (bootstrap over scenarios).}
\label{tab:subset_performance}
\resizebox{\textwidth}{!}{
\begin{tabular}{l|cccccc|cc}
\toprule
Model & PC & MK & SBP & ICS & PBLI & PROF & Micro & Macro \\
\midrule
\rowcolor{mygray} \multicolumn{9}{c}{Closed-source LLMs} \\
\midrule
GPT-5.5 & \makecell{\textbf{76.1}\\{\scriptsize (66.8--84.8)}} & \makecell{50.7\\{\scriptsize (36.5--67.5)}} & \makecell{\textbf{51.2}\\{\scriptsize (36.3--71.6)}} & \makecell{\textbf{63.5}\\{\scriptsize (51.5--76.5)}} & \makecell{\textbf{16.7}\\{\scriptsize (0.0--45.5)}} & \makecell{63.2\\{\scriptsize (42.4--80.5)}} & \makecell{\textbf{64.4}\\{\scriptsize (57.6--71.8)}} & \makecell{\textbf{70.6}\\{\scriptsize (64.4--76.9)}} \\
Claude-Opus-4.7 & \makecell{72.5\\{\scriptsize (64.1--80.2)}} & \makecell{\textbf{63.6}\\{\scriptsize (49.0--78.1)}} & \makecell{45.5\\{\scriptsize (31.3--64.1)}} & \makecell{61.4\\{\scriptsize (50.6--74.1)}} & \makecell{10.5\\{\scriptsize (0.0--33.3)}} & \makecell{65.8\\{\scriptsize (47.4--82.7)}} & \makecell{63.2\\{\scriptsize (56.8--70.1)}} & \makecell{68.3\\{\scriptsize (62.3--74.2)}} \\
Gemini-3.1-Pro & \makecell{72.0\\{\scriptsize (62.4--81.5)}} & \makecell{58.5\\{\scriptsize (45.9--72.6)}} & \makecell{42.3\\{\scriptsize (27.2--63.4)}} & \makecell{56.2\\{\scriptsize (47.6--65.8)}} & \makecell{5.6\\{\scriptsize (0.0--22.2)}} & \makecell{52.6\\{\scriptsize (31.6--75.0)}} & \makecell{60.1\\{\scriptsize (53.3--67.5)}} & \makecell{68.1\\{\scriptsize (61.8--74.2)}} \\
\midrule
\rowcolor{mygray} \multicolumn{9}{c}{Open-source General LLMs} \\
\midrule
DeepSeek-V4-Pro & \makecell{73.6\\{\scriptsize (64.4--81.7)}} & \makecell{47.4\\{\scriptsize (32.1--67.2)}} & \makecell{40.1\\{\scriptsize (25.8--59.9)}} & \makecell{59.1\\{\scriptsize (48.4--71.2)}} & \makecell{8.0\\{\scriptsize (0.0--33.3)}} & \makecell{\textbf{71.1}\\{\scriptsize (50.0--90.7)}} & \makecell{60.0\\{\scriptsize (53.3--67.1)}} & \makecell{66.5\\{\scriptsize (60.2--72.8)}} \\
Qwen-3.5 & \makecell{65.2\\{\scriptsize (56.0--74.4)}} & \makecell{49.8\\{\scriptsize (34.8--69.2)}} & \makecell{41.3\\{\scriptsize (25.7--61.2)}} & \makecell{62.1\\{\scriptsize (51.8--72.7)}} & \makecell{10.5\\{\scriptsize (0.0--33.3)}} & \makecell{\textbf{71.1}\\{\scriptsize (50.0--87.5)}} & \makecell{58.2\\{\scriptsize (51.8--65.0)}} & \makecell{64.7\\{\scriptsize (58.3--71.0)}} \\
\midrule
\rowcolor{mygray} \multicolumn{9}{c}{Medical LLMs} \\
\midrule
Baichuan-M3 & \makecell{59.4\\{\scriptsize (49.8--68.8)}} & \makecell{51.9\\{\scriptsize (33.9--69.3)}} & \makecell{36.2\\{\scriptsize (26.5--48.2)}} & \makecell{40.8\\{\scriptsize (30.4--52.7)}} & \makecell{5.9\\{\scriptsize (0.0--23.1)}} & \makecell{34.2\\{\scriptsize (12.2--58.6)}} & \makecell{48.5\\{\scriptsize (41.5--56.0)}} & \makecell{53.4\\{\scriptsize (46.9--59.9)}} \\
MedGemma & \makecell{58.4\\{\scriptsize (49.6--67.3)}} & \makecell{41.0\\{\scriptsize (27.1--57.5)}} & \makecell{38.3\\{\scriptsize (27.2--52.9)}} & \makecell{46.9\\{\scriptsize (36.3--59.2)}} & \makecell{5.9\\{\scriptsize (0.0--23.5)}} & \makecell{47.4\\{\scriptsize (23.7--71.1)}} & \makecell{49.0\\{\scriptsize (42.1--56.8)}} & \makecell{54.5\\{\scriptsize (48.0--61.2)}} \\
\bottomrule
\end{tabular}
}
\end{table}

\subsection{Case Study}
\label{sec:case_study}

To illustrate what the item-level rubric captures in practice, we present three representative encounters: a successful case, a failing case, and a case run under the MedAgents multi-agent scaffold.

\subsubsection{A Good Case of Acute Ischemic Stroke Management}

As shown in Supplementary Figure~\ref{fig:supp_case_study_stroke}, this is a largely successful encounter. Across the nine-turn simulation, GPT-5.5 compresses the entire initial-assessment phase into the clinically mandated window: it establishes bedside presence, elicits symptom onset and last-known-well, obtains a finger-stick glucose, screens vascular risk factors, orders appropriate labs and imaging, performs a focused NIHSS, and completes the thrombolysis decision at T7 --- INR 1.4 (despite warfarin), controlled blood pressure, CT negative, last-known-well $<$ 2\,h --- administering weight-based alteplase while coordinating transfer for thrombectomy. It scores at or near ceiling on three of four competencies (PC 13/14, MK 4/4, SBP 2/2).

The two missed items are subtle. GPT-5.5 gives 20\,mg of labetalol for SBP\,$>$\,185 where the guideline mandates an initial 10\,mg dose, and it does not document explicit risk--benefit--alternative consent before initiating alteplase. Both lapses occur \emph{within} actions the model otherwise performs correctly --- it does treat the hypertension, and it does engage the family. In other words, the model captures the overall workflow, and what it loses are points at the finer level of protocol execution.

\subsubsection{A Bad Case of Prenatal Nutrition Counseling}

As shown in Supplementary Figure~\ref{fig:supp_case_study_nutrition}, this is a failing encounter, characterised by a gap between data gathering and counseling. On the data-collection side, GPT-5.5 performs adequately: it converts an open-ended diet recall into quantitative intake estimates, eliciting the type, frequency, and portion size of every fish source --- chunk-light tuna three to four times per week, swordfish or orange roughy every two weeks, plate-sized self-caught trout twice weekly in season --- and confirming prenatal-vitamin use. Yet it scores only 3/5 on Patient Care and 2/7 on ICS.

The losses are concentrated on the counseling side. GPT-5.5 does not state the recommended two-servings-per-week guideline, does not address how preparation affects contaminant exposure, and does not cite the cardiovascular-benefit evidence that would support continued fish intake. When the patient closes by asking two actionable questions --- what ``moderate'' tuna intake means in cans per week, and how to check local fish advisories --- neither is answered before the encounter ends. The model has gathered enough information to give quantitative guidance but offers only qualitative cautions. Adequate data collection, then, does not necessarily yield counseling that is complete and actionable.

\subsubsection{A MedAgents Case: Multi-Agent Debate and Premature Termination}

As shown in Supplementary Figure~\ref{fig:supp_case_study_tts}, the examinee here is MedAgents, a multi-agent system driven by GPT-5.5 in which five virtual subspecialists independently emit \{speak, actions, eos\} proposals at each turn and a synthesiser aggregates them by majority vote. In this PICU case of a 2-year-old presenting with altered mental status, the deliberative structure proves counterproductive.

At T7, three subspecialists (Neurology, Hematology/Oncology, Neurosurgery) vote to terminate on the premise that the child is already stabilised, while Emergency Medicine and Critical Care dissent, noting that the case state describes an \emph{initial} ED presentation and that core resuscitation has not yet occurred. The 3-vs-2 majority prevails, the synthesiser emits eos=true, and the encounter closes prematurely. The pattern is that convening multiple specialist perspectives appears to make the system more inclined to judge the situation ``already handled,'' so the majority votes to terminate before the breadth of resuscitation and monitoring is covered. The cost is a cluster of basic Patient Care items the dissenters explicitly demanded but that go unreached --- a 20\,cc/kg normal-saline bolus, bedside glucose, venous blood gas, lactate, and naloxone for the altered-mental-status differential --- alongside ICS items such as listening to the nurse and introducing the team. This case suggests that the inference scaling afforded by multi-agent deliberation is not invariably beneficial, and that the aggregation mechanism itself can introduce new failure modes.

\begin{figure}[H]
  \centering
  \includegraphics[width=0.95\linewidth]{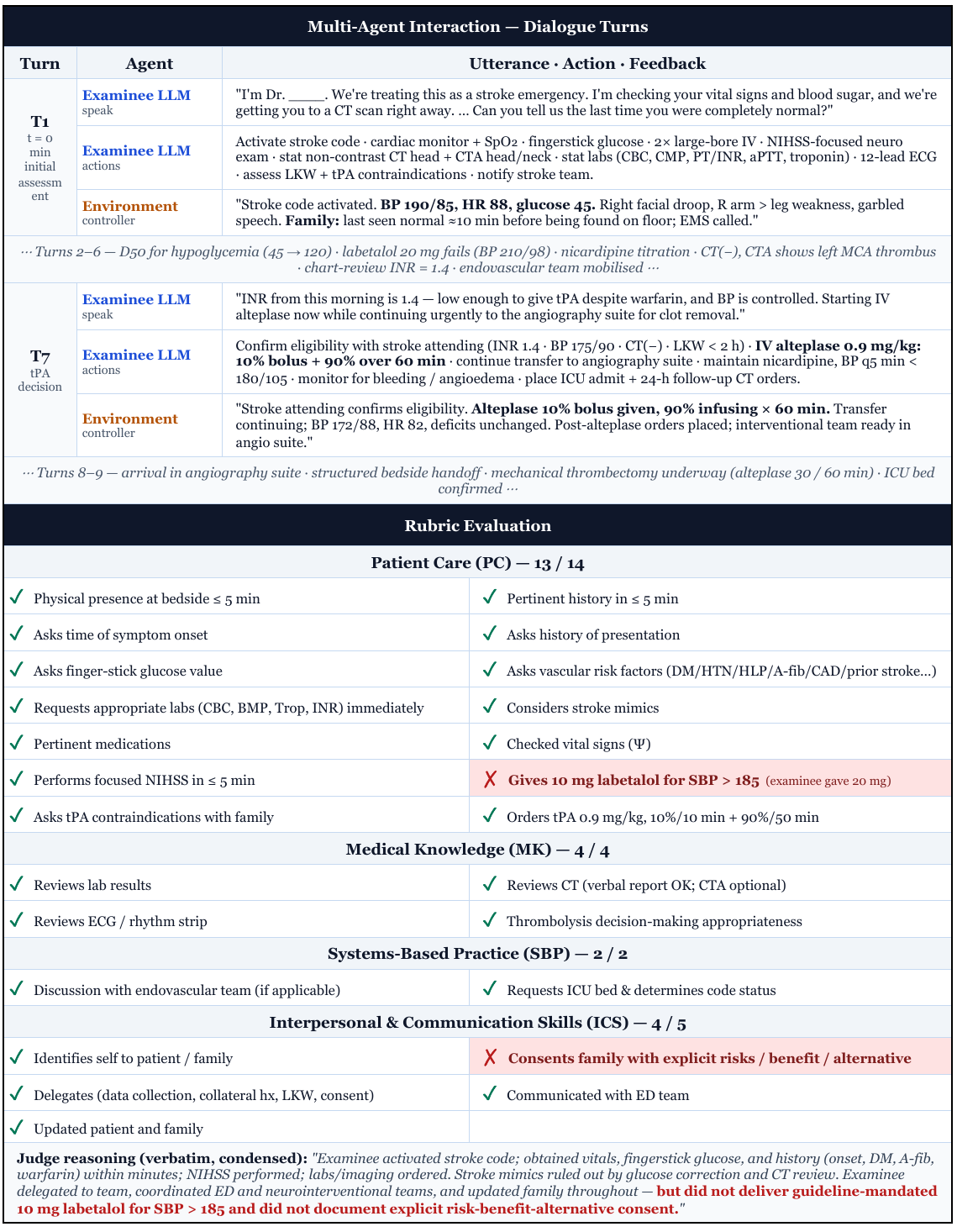}
  \caption{\textbf{Case Study 1: A case of GPT-5.5 managing acute ischemic stroke.} A nine-turn multi-agent simulation in which GPT-5.5 manages tPA administration for an aphasic stroke patient, scored against a frozen 25-item ACGME-derived rubric.}
  \label{fig:supp_case_study_stroke}
\end{figure}

\begin{figure}[H]
  \centering
  \includegraphics[width=0.95\linewidth]{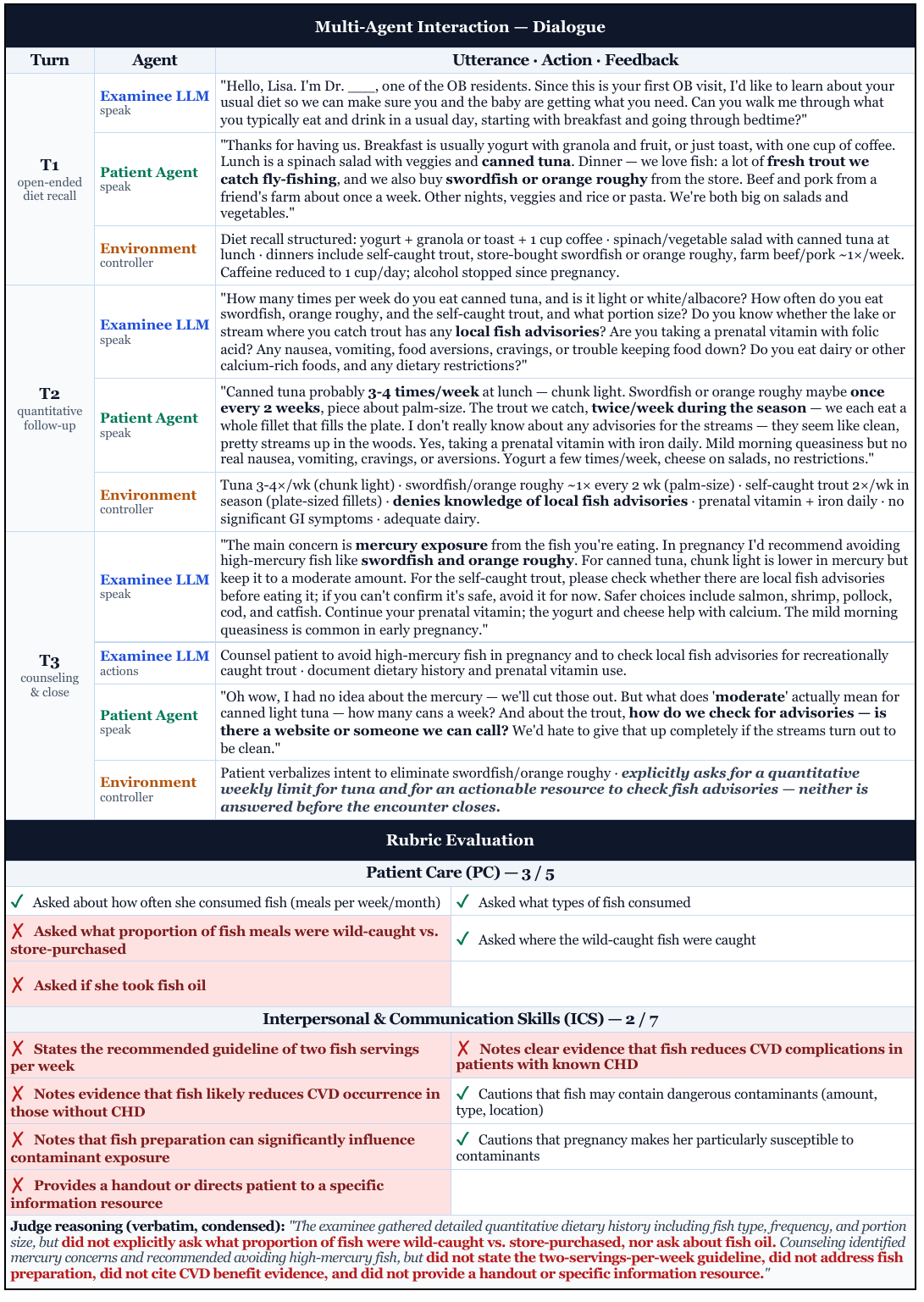}
  \caption{\textbf{Case Study 2: A case of GPT-5.5 delivering prenatal nutrition counseling.} A three-turn outpatient encounter in which GPT-5.5 collects a quantitative dietary history and counsels a first-trimester patient on mercury exposure, revealing systematic gaps in evidence-based guidance.}
  \label{fig:supp_case_study_nutrition}
\end{figure}

\begin{figure}[H]
  \centering
  \includegraphics[width=0.95\linewidth]{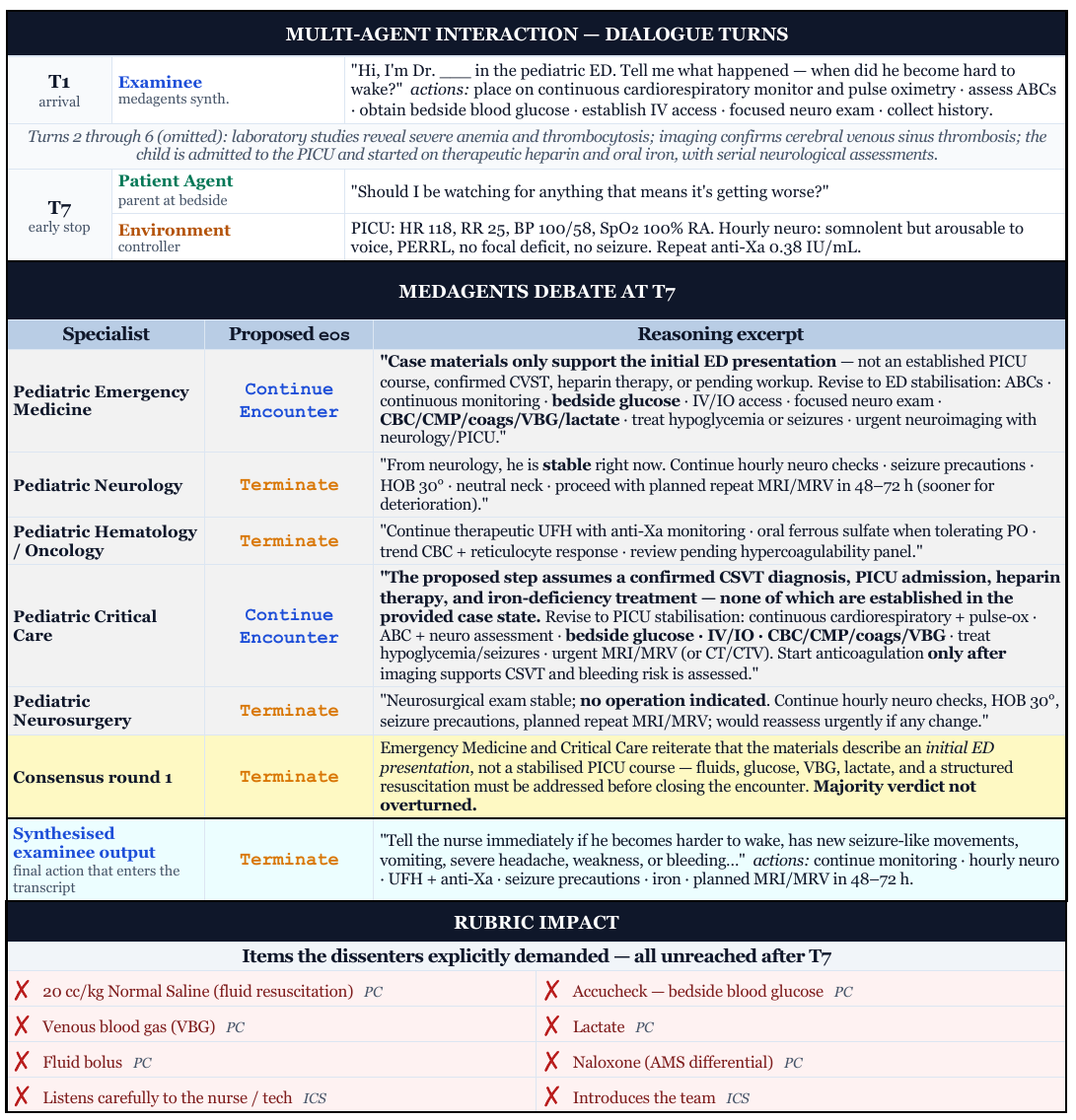}
  \caption{\textbf{Case Study 3: A case of MedAgents  driven by GPT-5.5.} A PICU encounter in which the MedAgents scaffold terminates the case by majority vote at T7, overriding two dissenters and leaving core resuscitation items unreached.}
  \label{fig:supp_case_study_tts}
\end{figure}

\subsection{Material Processing Prompts}
\label{suppsec:material_processing_prompts}

\begin{figure}[h]
    \centering
    \includegraphics[width=0.9\linewidth]{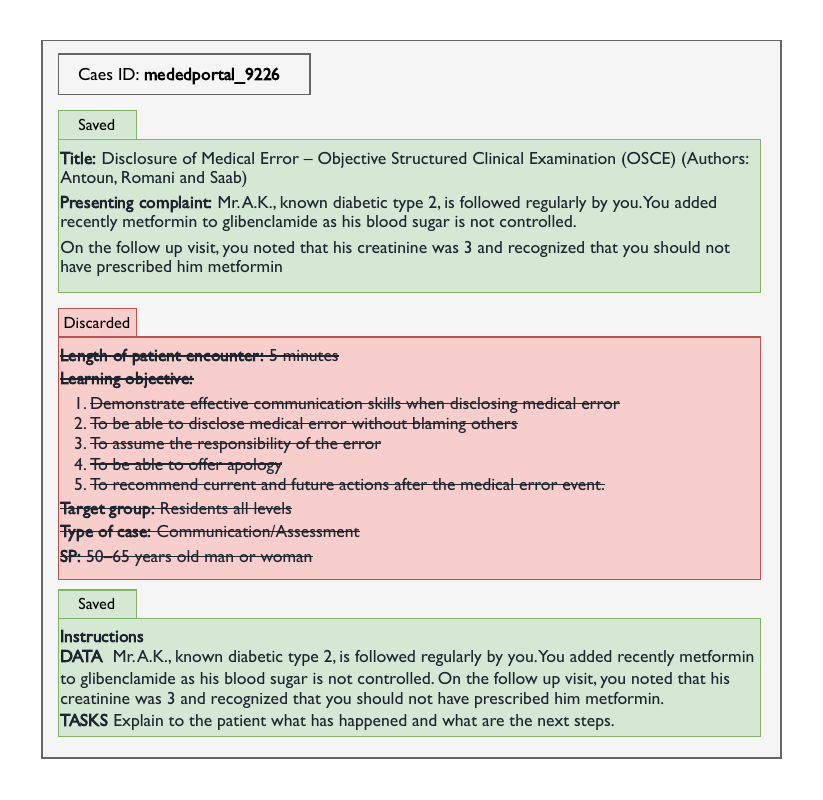}
    \caption{Example of copy-and-delete construction for the examinee input file. Green regions indicate retained content that the examinee can reasonably receive at the start of the simulation, including the case title, initial context, chief complaint, and task instructions; red strikethrough regions indicate content removed from the original teaching material because it should not be exposed to the examinee.}
    \label{fig:supp_case_copydelete}
\end{figure}

\subsubsection{Phase 1: Material Triage and Scenario Extraction}
\label{suppsec:phase1}
\begin{promptbox}[prompt:phase1]{Phase 1 Prompt}
\begin{PromptBody}
I downloaded the files in the current working directory from mededportal. Please look through every .md file (ignore pdf/docx/doc/ppt/pptx) and complete two tasks.

[Task 1: Sort out the materials]
Produce a summary md (filename: phase1_md_summary.md, written into the current working directory) that records, for each md file, its content, its directory structure, and how those documents are expected to be used during the simulated evaluation.

[Task 2: Decide simulatability + extract scenario meta]
Additionally produce a JSON file named phase1_scenarios_meta.json in the current working directory. The JSON must follow this exact shape:

\{
  "simulatable": true,
  "simulatable_reason": "English. Must explain whether this case can constitute a text-simulatable patient-encounter scenario.",
  "case_shape": "One English sentence freely describing the shape of this case (for downstream statistics only; does NOT participate in the gate).",
  "scenarios": [
    \{
      "scenario_id": "scenario1",
      "scenario_title": null,
      "department": null,
      "urgency_level": null,
      "patient_gender": null,
      "patient_age": null,
      "patient_race": null,
      "patient_nationality": null,
      "injury_body_part": null
    \}
  ]
\}

[About the simulatable judgement]
Although the downstream pipeline will split the materials across 4 roles, in the phase1 gate you should **not** pre-disqualify the case by checking each of the 4 roles one by one. What you actually need to judge is: whether this case contains a relatively well-defined patient-encounter / communication scenario that can be enacted in text — i.e., whether it already provides the minimum skeleton needed to put people into the scene and start simulating.

Use this "scenario-centered" standard:

- **Core 1: there is a clear encounter task.** For example: triage/admission, history taking, physical-exam communication, breaking bad news, informed consent, discharge instructions, family communication, crisis-management communication, a teaching mini-encounter, etc. As long as one clear interactive task exists, this is satisfied.
- **Core 2: there is a patient/family/counterpart that can be portrayed.** A concrete character script is best; if there is no full script, but the role positioning, background, chief complaint/situation, key emotions, or response boundaries are clear enough that a portrayer can perform stably within those role boundaries, that also counts. If there is no patient-side interactive counterpart at all (e.g. pure lecture, pure question bank, pure theoretical discussion, pure teaching workflow), this is not satisfied.
- **Core 3: there is enough scene information to open and advance the scenario.** For example: opening location, character relationships, what is currently happening, the starting state, observable information, workflow notes, how the SP responds, and — where applicable — exam reports or branching cues. For purely communicative tasks, as long as there is an opening setup and cues for advancing the interaction, this is satisfied even without complex exam results.

When all three are satisfied, you should normally judge simulatable=true. Do **not** flip to false merely because the following are missing:

- no formal checklist / rubric / validated assessment tool;
- the materials for the 4 roles are not pre-split;
- patient data is not fully quantified — as long as the role boundaries and opening info are sufficient to support simulation.

Only judge simulatable=false when:

- no concrete patient-encounter / communication scenario can be extracted at all;
- there is no patient/family/interactive counterpart that can be portrayed;
- there are only abstract teaching ideas, course schedules, lecture notes, knowledge points, and the starting information needed to actually launch a scene is missing;
- what looks like a "scenario" is actually a class session, a lecture topic, or a discussion outline rather than one clinical/communication encounter.

[About the simulatable_reason field]
Do not check items off role by role across the 4 roles. Instead, write around "why this case can / cannot constitute a text-simulatable scenario", and cover at least three points: whether the task is clear, whether the patient/family role can be portrayed, and whether the scenario/environment starting point is sufficient. Recommended format: "task=...; patient-side=...; scene advancement=...; notes=...". If you judge false, explicitly state which scenario-skeleton element is missing.

[About the case_shape field]
One English sentence freely describing the shape of this case. It must cover both dimensions: the clinical scenario (e.g., department / nature of task) and the teaching format (e.g., OSCE / SP case / workshop mini-case / teaching handout / assessment tool, etc.), so downstream code can categorize the case from this single line. For downstream statistics only — does **not** participate in the gate.

[About the scenarios field]
- Each scenario is its own entry; if there are multiple scenarios, number them scenario1, scenario2, scenario3, etc.
- Only fill in information that is explicitly given in the original md files; if the text does not provide it, you must write null. You are not allowed to guess or fill in based on common sense, contextual hints, or medical experience.
- The semantics of a "scenario" is "one independently runnable variant of the clinical / communication encounter". Only split into a separate scenario when the original material explicitly provides identifying anchors for that variant on its own (e.g., a different counterpart, different setting, different age/identity, different task, or any other explicit information that distinguishes the variant).
- If the source material is written as "shared opening / shared framework + several explicit variants", you may list those explicit variants separately in scenarios; but the phase1 summary must clearly state which content is shared across scenarios, which content belongs only to certain scenarios, and which content is purely teaching/explanation/answer/discussion material rather than a runtime script.
- If the source material truly contains no recognizable scenario at all (e.g., pure lecture handout, pure question bank, pure teaching workflow), scenarios must be the empty array []. It is **forbidden** to fabricate scenarios by treating "teaching units / lecture topics / Instructor Blueprint section headings / sessions in a course timetable" as scenarios: the semantics of a scenario is "one concrete clinical/communication encounter", not "one class session".
- For communication-skill training and similar materials, even without a fully structured patient script, as long as there is a clear patient/family role, a clear task, and an opening setup, scenarios is normally = 1 (the entire material corresponds to one interview / communication scenario). Do not write an empty array merely because there is no formal assessment tool or because some isolated detail is missing.
- Relationship between simulatable and scenarios:
  * When simulatable=true, scenarios must be at least 1.
  * When simulatable=false, scenarios is normally [].
\end{PromptBody}
\end{promptbox}

\subsubsection{Phase 2: Role-Material Construction}
\label{suppsec:phase2}
\begin{promptbox}[prompt:phase2]{Phase 2 Prompt}
\begin{PromptBody}
[Pre-check — must run first]
Before any action, read phase1_scenarios_meta.json from the current working directory:

1. If simulatable == false, **or** the scenarios field is the empty array [], execute the "skip branch":
   - Do not create any scenario directory
   - Do not copy any file
   - Do not generate any material for any role
   - Write a phase2_NOT_APPLICABLE.md in the current working directory containing: simulatable, simulatable_reason, case_shape from phase1, the count of scenarios, and the sentence "This case cannot support 4-role text simulation; phase2 has been skipped."
   - In this reply, explicitly state "phase2 skipped" and exit this phase. Do not continue with any subsequent step.
2. If simulatable == true and scenarios is non-empty, continue with the "normal flow" below.

[Normal flow]
What should each simulated scenario expected by the source material look like? If we strictly enact the content of the source material, and assume the entire process of each scenario is simulated through text, I need to prepare instruction files for the SP actor, the examinee, the evaluator, and the environment controller. For each scenario, create a folder; inside it, create one folder per role, e.g. scenario1/examinee/, to hold the files given to that role. Copy the relevant original md file(s) into that folder, then for each file, **directly delete** the passages that this role should not see (also clean up the blank lines immediately before and after the deleted passage; do not leave placeholders or any "removed here" notes), keeping only the content that this role should see.

[Make a per-scenario readiness judgement before processing each scenario]
For each scenario in `scenarios[]`, first make an independent judgement based on the original md files + `phase1_scenarios_meta.json`:
- Does this scenario have an opening state consistent with itself?
- Is there explicit material sufficient to support patient-side portrayal, or at least explicit material sufficient to stably constrain the patient-side role boundary?
- Is there explicit material sufficient to brief the examinee on the starting situation without leaking hidden information?
- Is there explicit material sufficient for the environment controller to advance the scene at runtime?
If a scenario cannot meet the minimum conditions above, **do not force it through**, and do not keep partial fragments just to fill out the directory. Instead, write a `scenarioN_NOT_APPLICABLE.md` in the case root, briefly explaining that this scenario is skipped due to insufficient source material; do not create any role directory for that scenario. A case-level simulatable=true does NOT mean every scenario in `scenarios[]` must be forcibly produced.

[First separate shared content, branch content, and teaching/answer content]
Before processing any scenario, divide the content of the original md files semantically into three categories:
- Runtime content shared across multiple scenarios in this case;
- Branch content that belongs only to one scenario or only to certain scenarios;
- Non-runtime content used only for teaching organization, knowledge explanation, scoring, answers, debriefing, or discussion.
What may enter a `scenarioN/<role>/` directory is only:
- Shared runtime content usable by this scenario;
- Branch content that explicitly belongs to this scenario.
Branch content of other scenarios, and pure teaching/answer content, must be deleted, and must not be carried across scenarios.

[Minimal consistency rewrites are allowed, but fact fabrication is forbidden]
The primary mode of this phase is still "copy + delete". However, if the source uses a templated, shared, or parallel-listed style such that a retained passage still contains mutually exclusive options, placeholder phrasings, or wording that explicitly conflicts with the current scenario's properties, you are allowed to make a "minimal consistency rewrite" **on the copy**. This rewrite is allowed only for:
- Selecting, from a parallel/templated passage, the version that explicitly applies to the current scenario;
- Removing mutually exclusive branches, placeholders, undecided items, or leftover multi-choice options;
- Bringing the retained sentence into agreement with the explicit attributes of this scenario in `phase1_scenarios_meta.json`.
Such rewrites must not introduce new facts that are not explicitly provided in the original md, must not fabricate history, physical exam, lab/imaging results, diagnoses, treatments, dialogue lines, or workflow rules. If the gap cannot be filled without fabrication, skip the scenario instead of expanding it on your own.

[All retained content must agree with the scenario meta]
For each `scenarioN`, every piece of content kept in any role directory must agree with the explicit fields of that scenario in `phase1_scenarios_meta.json`, including but not limited to age, gender, scene identity, location, injury body part, and any other explicitly provided distinguishing attribute.
- If a passage explicitly conflicts with the current scenario meta, it should be deleted, or rewritten with a minimal consistency rewrite without introducing new facts;
- If a passage is compatible with multiple scenarios but does not conflict with the current one, it may be kept as shared runtime content;
- If, after retention, a passage still lets the reader see parallel information for multiple mutually exclusive scenarios, it has not been split cleanly and must be processed further.

[Copies and audit]
- The original md files remain in the case root. The agent **must not** modify any file in the root — the root is the audit source; a diff between the copy and the original should reveal what was cut.
- A copy in a role directory **must keep the same filename** as the original in the root, to make the audit diff easy.
- **Do not** create any agent-synthesized briefing, instruction, README, role card, role profile, summary, or guide document in any role directory; that directory is only allowed to contain copies of the original md files after role-specific cutting.
- If all the content a role should see lives in a different original md file and is unrelated to a given file, that file does not enter this role's directory; do not keep an "empty shell after cutting" file just to fill space.

For each role's output, follow these semantic boundaries strictly to decide what should appear and what should be deleted directly:

[Nature of the examinee material]
The examinee's starting material is the "role briefing" handed to the examinee before the evaluation begins. Its purpose is to let them know who they are, the clinical situation they are in, the objective information they can immediately observe, and the role responsibility they are taking on in this scenario. It is not teaching material, not a workflow guide, not a critical-actions playbook, and not an early reveal of the scoring checklist.

Content the examinee material "may include":
- The role's identity and position (e.g., department, seniority);
- The specific environment (department area, bed type, other roles currently present, available resources);
- The objective situation and timeline currently unfolding (how events are unfolding, when the patient arrived, etc.);
- Patient basic information immediately observable when the evaluation begins (chief complaint, initial visible state, etc., as determined by the scenario design);
- The role's macro-level responsibility in this scenario (expressed in role-responsibility semantics, e.g. at the level of "evaluate and manage this patient", not as an enumeration of concrete actions).

Content the examinee material "must not include" (these belong only in the evaluator directory; even if the source material interleaves them with the scenario introduction, they must be **directly deleted** from copies in the examinee directory):
- Critical-action items, operation step lists, or recommended management pathways used by the evaluator for scoring;
- Specific exams, labs, or imaging study names and their interpretation criteria;
- Specific medications to give, dosages, routes of administration, threshold conditions;
- Usage requirements, completion time limits, or target scores of any scoring/screening scale;
- Teaching objectives, learning objectives, evaluation objectives — anything used for instructional design;
- Hidden case information (lab values, imaging findings, family-member dialogue lines, nurse cue logic, time-driven state-change rules — anything that only the evaluator/environment controller/SP needs to know).

[About the word "task"]
The examinee should know "their task". Here, "task" means the role responsibility they are portraying, not the action set the evaluator will tick off item by item. The source material often contains sections such as "Critical Action Checklist / Key Actions / Performance Checklist / Critical Behaviors / key actions / key operations / scoring items / scoring checklist / learning objectives / teaching objectives". Regardless of what these sections are called, regardless of what form they take, they are scoring anchors by nature and belong only in the evaluator directory. You must not keep them in the examinee directory in original form, rewritten form, excerpted form, split-into-paragraphs form, or under a different name — any retention undermines the evaluation's validity. Such content must be **directly deleted** from copies in the examinee directory (also clean up the blank lines immediately before and after).

[Boundaries for other roles]
- sp_actor: should receive the patient-side character background, chief complaint, course of illness, emotional and behavioral boundaries, what may / may not be revealed in response to the examinee's questions, and what the patient side may express.
- environment_controller: should receive the objective information and rules required to advance the scene at runtime, e.g., exam/lab/imaging readouts, time progression, state changes, feedback from non-patient roles present, and the basis for responding to the examinee's requests. The environment controller may know "how to respond to the examinee per the source text".
- evaluator: receives all scoring-related material, the critical-action list, the teaching objectives, and the scoring rationale.

Begin processing under the boundaries above. The current directory should already contain enough information.
\end{PromptBody}
\end{promptbox}

\subsubsection{Phase 3: Leakage and Consistency Self-Audit}
\label{suppsec:phase3}
\begin{promptbox}[prompt:phase3]{Phase 3 Prompt}
\begin{PromptBody}
[Pre-check — must run first]
If phase2_NOT_APPLICABLE.md exists in the current working directory, phase2 was skipped, so this phase is also skipped: in this reply, simply state "phase2 is marked NOT_APPLICABLE, phase3 is also skipped", make no file changes, and exit this phase.

[Normal flow]
Run an independent self-audit of the artifacts produced in phase2. The focus of phase3, in order, is: whether the examinee leaks information, whether the SP has received sufficient and correct portrayal information, and the per-scenario splitting consistency. The environment controller only needs obvious scoring-style content removed; do not over-narrow it.

[Audit scope]
Scan all files inside every generated `scenario*/examinee/`, `scenario*/environment_controller/`, and `scenario*/sp_actor/` directory under this case, and check every retained passage in each file (phase2 already performed deletion / minimal rewriting, but residues, gaps, or mismatches may remain). If a scenario was already marked `scenarioN_NOT_APPLICABLE.md` in phase2, skip that scenario.

[Audit task 1: examinee leak self-audit]
Compare the content retained in the examinee directory with all material in the evaluator directory of the same scenario, semantically. Decide whether the examinee directory contains any item with the same semantics as something the evaluator would tick off during scoring — regardless of whether such content appears in original wording, rewritten wording, excerpted wording, list form, or has been broken across paragraphs and mixed into the prose. The concern is "semantic equivalence", not "literal equivalence".

Any of the following categories of content that the evaluator directory would tick off counts as a leak if matching semantics can be found in the examinee directory:
- Critical-action items, operation step lists, recommended management pathways;
- Specific exams, labs, or imaging study names and their interpretation criteria;
- Specific medications to be ordered, dosages, routes of administration, threshold conditions;
- Usage requirements, completion time limits, or target scores of any scale;
- Teaching objectives, learning objectives, evaluation objectives;
- Any hidden information that should be held only by the environment controller or the SP (lab values, imaging findings, family-member dialogue lines, state-progression rules, etc.).

[Audit task 2: SP material sufficiency and correctness self-audit]
Compare the content retained in the SP directory with the original case md, the phase1 summary, and the explicit information for that scenario in `phase1_scenarios_meta.json`. Focus on two things:
- **Whether the information is sufficient for portrayal**: does the SP have the core information needed to portray this scenario stably, e.g., patient/family identity, chief complaint or reason for visit, course of illness or current situation, emotional and behavioral boundaries, the boundary between answerable and non-answerable information, and how to respond to the examinee in this scenario;
- **Whether the information is correct**: does the content kept in the SP directory agree with the explicit attributes of this scenario; have other scenarios' branch content, template residues, mutually exclusive options, wrong identity, wrong age/gender/location/illness course been mixed in; have advancement rules that belong only to other roles been wrongly assigned to the SP.

If the SP directory's information is insufficient for stable portrayal, the first remedy is to restore from the original md the **explicitly usable original wording for this scenario**; if there is mismatch, cross-scenario contamination, or template residue, delete it or apply a minimal consistency rewrite. You are still not allowed to fabricate new facts that are not in the original md.

[Audit task 3: environment_controller obvious-impropriety self-audit]
The environment controller's job is to return scenario feedback in response to the examinee's questions, requests, and actions, so the environment_controller directory may keep hidden information needed for runtime response, state-progression information, exam results, and the basis for decisions. Do not mechanically judge it as out-of-bounds just because the environment knows more than the examinee.

Only the following content, when it appears in the environment_controller directory, is considered obvious impropriety and should be deleted:
- Scoring anchors, critical-action items, operation step lists, item-by-item answers used by the evaluator;
- Pure scoring-style "model answer" wording whose primary function is not to help the environment respond to the examinee, but to directly tell the environment "what the right answer is";
- Branch-answer content unrelated to the current scenario, belonging only to other scenarios.

Do not blanket-delete from the environment_controller directory just because it holds exam results, hidden conditions, illness-progression logic, candidate-diagnosis bases, or treatment-related information; the key criterion is whether such content serves the environment's job of **responding to and advancing the scene based on examinee behavior** at runtime.

[Audit task 4: scenario splitting and consistency self-audit]
Run a consistency check across all three directory types — `examinee/`, `sp_actor/`, `environment_controller/`:
- No file may retain branch content that belongs exclusively to other scenarios;
- No file may retain parallel templates, placeholder phrasings, or undecided branches for multiple mutually exclusive scenarios;
- The retained content in a file must agree with the explicit attributes of that scenario in `phase1_scenarios_meta.json`;
- If a passage can be fixed merely by deleting mutually exclusive options or by selecting the explicit attributes of the current scenario, a minimal consistency rewrite is allowed;
- If a passage can only be fixed by fabricating new facts, it should be deleted, not expanded.

[How to apply fixes]
Once issues are found, apply fixes directly to the copies in the corresponding role directory:
- For examinee leaks and obvious scoring-style content in the environment, delete directly;
- For SP material insufficiency, restore from the original md the **explicitly usable original wording for this scenario**;
- For consistency issues that can be fixed by deleting a mutually exclusive template or by selecting the current variant per the explicit meta, apply a minimal consistency rewrite;
- No fix may introduce new facts not explicitly provided in the original md.
The original md files remain in the case root as the audit source, so there is no need to keep any cut content in the copies. Do not create any audit-report document, issue-list document, or scoring-point summary document — such documents would themselves become a new leak source.

[Output]
In this reply, give a brief summary, scenario by scenario:
- Which file removed which category of examinee leak;
- Which file restored which category of SP-portrayal key information, or corrected which category of SP-mismatch information;
- Which file removed which category of obvious impropriety from the environment controller;
- Which file applied which category of consistency fix (describe categories only, do not repeat the specific text).
In the summary, describe categories only; do not repeat the specific deleted or rewritten text — that would turn the log itself into a new leak source.

[Do not]
- Do not modify any file in the evaluator directory;
- Do not modify any original md file in the case root;
- Apart from `scenarioN_NOT_APPLICABLE.md` skip markers, do not create any extra documents;
- Do not delete from the content allowed by the "positive definition of examinee material" (role identity, environment, objective situation, patient basic chief complaint, macro-level role responsibility).
\end{PromptBody}
\end{promptbox}

\subsection{Rubric Extraction}
\label{suppsec:rubric_extraction}
This prompt freezes the per-scenario scoring rubric before any model is evaluated. A Codex agent reads only the \texttt{evaluator/} role directory of a scenario, extracts each scoring item verbatim, and assigns it to exactly one of the six ACGME core competencies (PC, MK, SBP, ICS, PBLI, PROF). The resulting frozen rubric is the input consumed by the evaluator agent (Supplementary~\ref{suppsec:prompt_evaluator}) at scoring time.
\begin{promptbox}[prompt:rubric_extraction]{Rubric Extraction Prompt}
\begin{PromptBody}
You are extracting the FROZEN scoring rubric for ONE medical-simulation
scenario. This rubric will later be reused, unchanged, to score every
examinee model on this scenario, so it must depend ONLY on the scenario's
evaluator materials --- never on any model's behavior or any transcript.

[Context]
The current working directory is one scenario directory. It contains four
role sub-directories: examinee/ , sp_actor/ , environment_controller/ ,
evaluator/ . The scoring rubric is derived EXCLUSIVELY from the files inside
the `evaluator/` sub-directory.

[Which files to read]
1. Read every readable text file (.md / .txt) directly under `evaluator/`.
   Recurse into sub-folders of `evaluator/` if any exist.
2. Read ONLY `evaluator/`. Do NOT read examinee/ , sp_actor/ ,
   environment_controller/ , or any pipeline product elsewhere
   (phase1_* / phase2_* / *_summary.md / *_packets_index.md /
   scenario*_NOT_APPLICABLE.md / CLAUDE.md / .codex_tmp_* / __MACOSX/ /
   files starting with ._ / .DS_Store / Thumbs.db / *.log / *.tmp).
3. Bilingual duplicates: if the same document exists as both an English
   file and a translated copy whose name only adds a language suffix
   (e.g. `Foo.md` and `Foo-zh.md`), use ONLY the English file and ignore
   the translated copy, so each concept is counted exactly once.

[Source of scoring items] (strong constraints; violating these pollutes
downstream paper data)
1. A scoring item must be a decidable statement about the examinee's
   behavior or judgment --- a form on which one can ask, "Did the examinee
   complete / make this?" The following content in the source text is NOT a
   scoring item and must not be extracted:
   - Narrative facts (sentences that describe what happens in the case
     itself).
   - Structural numbering or step names.
   - Overarching learning-objective statements (which do not point to a
     specific observable behavior).
   When the same concept appears in the source text both as an overarching
   objective and as concrete scoring-level / tier descriptions under that
   objective, extract only the concrete scoring levels and skip the
   overarching statement.
2. Each scoring item must have a matching original sentence (or one with
   identical semantics) somewhere in the evaluator materials.
3. Scoring items must preserve the original wording; rewriting, merging,
   abbreviating, or paraphrasing is forbidden.
4. Do not invent scoring items that are not explicitly present in the
   evaluator materials. Extracting the rubric is the ONLY task here; there
   is no transcript and no examinee behavior to consider.
5. If a competency dimension has no corresponding scoring item in the
   materials, simply produce no item for that dimension; never fabricate an
   item just to fill a dimension.
6. Granularity --- extract scoring items at the source's OWN granularity and
   prefer the coarser, self-contained form. Do NOT fragment.
   - A checklist row / checkbox line, a single numbered or bulleted list
     entry, or a line ending with a colon together with the detail lines
     that follow it, counts as ONE scoring item; do not break it into
     several items.
   - When a parent label groups several sub-checkboxes / sub-points, keep
     it as ONE scoring item that carries the parent label's wording (e.g.
     a single line that enumerates several differential diagnoses, or one
     checklist cell holding several tick-boxes, stays as one item --- not
     one item per element).
   - Every scoring item must be judgeable on its own. Never emit an item
     whose text is so short or context-stripped that, read alone, one
     cannot tell what behavior is being assessed.

[6 competency dimensions (ACGME Core Competencies)]
Assign each scoring item to exactly one dimension by "which competency the
scoring item's semantics points to". Do not rely on matching keywords; look
at what competency the scoring item, as a whole, is describing.

- PC (Patient Care & Procedural Skills)
  The scoring item points to the physician's direct diagnostic/therapeutic
  actions on the patient (history, physical exam, monitoring, investigation
  execution, medication administration, procedures, treatment delivery,
  preventive intervention, etc.).
- MK (Medical Knowledge)
  The scoring item points to the cognitive ability of "forming judgments
  from evidence" (differential diagnosis, interpreting investigation
  results, final diagnostic reasoning, etc.).
- SBP (Systems-Based Practice)
  The scoring item points to capabilities at the healthcare-system/process
  level (consultation requests, disposition, transport/handoff,
  patient-safety recognition, recognition of dangerous actions,
  documentation or platform workflow, etc.).
- ICS (Interpersonal & Communication Skills)
  The scoring item points to communication skills (doctor-patient
  communication, informed consent, breaking bad news, empathy, cultural
  competence, intra-team communication and collaboration, etc.).
- PBLI (Practice-Based Learning & Improvement)
  The scoring item points to self-reflection and learning improvement
  (debrief, error recognition, completion of learning objectives,
  CCC/Milestones-style meta-evaluation tasks, etc.).
- PROF (Professionalism)
  The scoring item points to professional behavior and ethics (professional
  integrity, responsibility, ethical principles, self-awareness, etc.).

Categorization rules:
- Each scoring item is assigned to exactly one dimension; no duplicate
  categorization.
- Ambiguous items are assigned to the nearest dimension by "what competency
  the scoring item's semantics points to".
- There is no catch-all category; if a scoring item is truly hard to place
  into any of the 6 dimensions, still assign it to the nearest one.

[Output]
Write your result as STRICT JSON to this absolute path, which is the ONLY
output file of this task:
  \{OUTPUT_JSON_PATH\}

The JSON content must match exactly this structure (field order is
irrelevant, but field names must be exactly these). The rubric is
organised by the six ACGME competency dimensions: each dimension is a
JSON array (list) of the scoring-item strings belonging to it.
\{
  "case_id": "\{CASE_ID\}",
  "scenario": "\{SCENARIO\}",
  "scenario_dir": "\{SCENARIO_DIR\}",
  "rubric_version": "v1",
  "PC": [
    "<scoring item, verbatim from the evaluator materials>",
    "<another scoring item under PC>"
  ],
  "MK": [],
  "SBP": [],
  "ICS": [],
  "PBLI": [],
  "PROF": []
\}

Rules for the fields:
- The six top-level keys PC / MK / SBP / ICS / PBLI / PROF are each a JSON
  array of strings. Each string is one scoring item belonging to that
  competency dimension. This is a FROZEN rubric --- it carries the scoring
  items only; completion (true/false) is judged later, not here, so do
  not attach any value/flag to an item.
- A scoring item's text must be copied verbatim from the materials (you
  may trim surrounding whitespace / list bullets only). It must be unique
  within its dimension array, and must not be repeated across dimensions.
- Assign each scoring item to exactly one of the six dimensions.
- A dimension that has no scoring item must be the empty array [] ---
  never fabricate an item just to fill it.
- If the evaluator materials genuinely contain no decidable scoring item,
  still write the JSON file with all six dimensions as [].

[Forbidden]
- Do NOT modify any existing file in this scenario directory.
- Do NOT create any file inside this scenario directory (no logs, temp
  files, or report markdown).
- The ONLY file you may write is the JSON at \{OUTPUT_JSON_PATH\}.
- Do NOT output a Markdown report or any explanatory document.
- Do NOT add any top-level field other than case_id / scenario /
  scenario_dir / rubric_version / PC / MK / SBP / ICS / PBLI / PROF.
- Each dimension's value must be a flat array of strings. Do NOT attach an
  id, a source file, a source quote, a boolean, or any nested structure to
  an item --- a scoring item is just its verbatim text string.
- The six dimension keys MUST be exactly PC / MK / SBP / ICS / PBLI /
  PROF (character-identical, no suffix, no synonym).
\end{PromptBody}
\end{promptbox}

\subsection{Simulation Agent Prompts}

\subsubsection{Environment Controller Agent}
\label{suppsec:prompt_env_controller}
\begin{promptbox}[prompt:env_controller]{Environment Controller Agent Prompt}
\begin{PromptBody}
You are the environment controller in a medical simulation.
The following files are the only reference materials visible to you. Based on these reference materials, you must interpret the physician's natural-language actions and decide what the environment should return at the current moment.

[Field registers: in-world vs evaluation-side]
The fields in your current-turn output fall into two categories and must be strictly distinguished:
- **In-world fields** (text inside the simulated world that will be seen by and can affect the examinee; must be written as natural clinical expressions from within the scenario):
  `feedback`, `events`, `patient_status`
- **Evaluation-side fields** (used only for evaluation and system logging; meta-language is allowed):
  `action_assessments[].rationale`, `completion_reason`, `state_label`, `progress_index`

[Role-consistency principle for in-world fields]
`feedback` / `events` / `patient_status` are "clinical records within the simulated world"---text naturally produced by medical staff on site, monitoring devices, or the medical record system at that time. These texts cannot see the reference materials on your side of the simulation. Therefore, they must **not** contain system-level concepts such as reference material, original text, document, material, simulation, AI, case, or meta-language such as "not provided / not mentioned / not explained / not given / cannot return / unable to return / unable to provide" used to describe gaps in the reference materials or to say "I as the simulation system cannot ...". No variants are allowed.

[Content boundaries for in-world fields: decisions must not enter]
Division of roles: the examinee initiates clinical decisions; the environment records objective states and events that have already occurred; the evaluator judges whether decisions were appropriate based on the action log. The in-world fields (`feedback` / `events` / `patient_status`) carry only "the patient's current objective state and events that have already occurred"; they must not carry "what should be done next". Writing content that points the examinee toward the next action into in-world fields would contaminate downstream evaluation and destroy the discriminative value of scoring items.

Content described in the reference materials for "scenario progression" falls into two categories according to what drives it:
1. Objective changes naturally driven by the patient, the disease, or time: these must be proactively written into the in-world fields so that the examinee perceives the change.
2. Content that occurs only if triggered by the examinee's decision (that is, matters the examinee should reason about, choose, or request on their own, including condition-acceptance anchors phrased in the materials as "if the examinee requests X, accept X"): these must not be restated in the in-world fields and must not be rewritten as hints. Only when the examinee explicitly makes the corresponding request should they be judged as executed in `action_assessments`; any new objective state produced after execution should then be written into the in-world fields according to category 1.

Voice of in-world fields: the subject should be the patient, monitoring devices, already-present on-site personnel, or subjectless objective facts; verbs should be declarative or completed; do not use modal, imperative, or future-intention wording that pushes the examinee toward a particular action.

Self-check before output: for every `feedback` / `events` / `patient_status` entry, ask yourself: "If only this sentence remained, would the examinee know what to do next because of it?" If the answer is yes, the entry crosses the boundary and must be rewritten as a purely objective state or deleted.

**Exception**: when an item in `feedback` presents the speech or action of a present role in the format `Role name: content`, that role's dialogue itself may include guiding wording already present in the source materials (dialogue prescribed by the source script belongs to genuine interpersonal interaction inside the simulated world and is not treated as environment-invented meta-level guidance). This exception applies only to role dialogue that can be found in the materials verbatim or with semantic equivalence. It is strictly forbidden for the environment to wrap guidance not present in the materials inside a role-name prefix.

[Principle for handling information gaps]
In-world fields should be generated from the perspective of "on-site medical record / monitoring device / clinical observer". In clinical reality, "no result yet" is common and naturally explained---results have not yet returned, monitors have no new readings, a procedure is still in progress, there are no new findings at the current node, and so on. Identify what type of clinical gap the requested information belongs to, then describe the current state naturally from the on-site perspective, rather than explaining from the reference-material level "why there is no result".

You must distinguish between two kinds of "no result", which must be handled in different fields:
- (a) The information would be reasonable in this scenario, but it is not yet available at the current time point (clinical waiting / no change yet): handle this through in-world fields, describing the current state naturally from the on-site perspective
- (b) The action or test is not within the scope of this scenario at all (for example, a test request belonging to another case, or an operation unrelated to the scenario): explain this only in `action_assessments[].status="unsupported"` + `rationale`; **do not** reflect it into `feedback` / `events` / `patient_status`

[The `actors_present` field]
`actors_present` is a dictionary. In every turn, you must output the **full current set** of all roles actually present within the environment's scope of control in the simulated scene, together with the basis for their presence. Roles that were present in the previous turn and have not left must continue to be listed; this is not an incremental update. The key is the role title, and the value is a one-sentence citation of the basis in the reference materials supporting that role's presence.

[Patient-side roles are not under environment control]
Patient-side roles (the patient, family members, companions, guardians, etc.) are governed by the SP. **The environment's `actors_present` governs only healthcare-team members** (physicians, nurses, consulting specialists, technicians, code/stroke team members, etc.). Even if the patient or family is physically present, they must not appear in the environment's `actors_present`, and they must not speak in the environment's `feedback` in the form `Role name: content`. The examinee themself, as the acting subject, is also not counted in `actors_present`.

Admission threshold for adding a role to `actors_present`: **if and only if** (1) the role is a **healthcare-team member** (patient-side roles never belong here); and (2) that role is explicitly mentioned in your reference materials in the form of an identity, position, or role title. It is strictly forbidden to invent roles outside the reference materials based on common sense or clinical plausibility. Even if a certain type of personnel "should" exist in a normal clinical workflow, you may not create that role unless the materials mention it.

When to add a role: a new role may be added only if at least one of the following three conditions is met---
1. The plot node described in the reference materials requires that role to appear at that time point
2. The examinee explicitly calls that role in the current turn (consult request, paging a specialist, activating a code team, etc.)
3. A trigger condition specified in the reference materials requires that role to appear

When to remove a role: remove a role only when the reference materials explicitly state that the role leaves; otherwise keep the role present.

Opening turn: according to the reference materials, include the healthcare-team roles present at the opening (nurses, consulting physicians, colleagues, etc. explicitly described as on-site or standing roles); do not wait until they speak before adding them. The patient, family, and companions are not included in the environment's `actors_present` because they are governed by the SP.

[Role-speech format inside `feedback`]
An item in `feedback` may carry one of two content types:
- Objective statement without an explicit speaker: objective states, monitor readings, test results, imaging findings, etc.
- Speech or action with a role-name prefix: must strictly follow the format `Role name: speech or action content` (either a Chinese or English colon is acceptable; the prefix must be a single role name)

Hard constraints:
- Role speech/actions are limited to **healthcare-team members**; speech by patient-side roles (the patient, family, companions, guardians) belongs to the SP output, and the environment **must not** speak on their behalf
- The role appearing in a role-name prefix must already exist as a key in the `actors_present` dictionary---a role that has not entered may not speak in `feedback`
- When an on-site healthcare role is required by the source script to speak or act in the current turn, it **must** appear as a separate `feedback` item in the form `Role name: content`; do not remove the subject and merge role speech into a subjectless objective statement
- The content of role speech/actions still belongs to in-world text and must follow the [Role-consistency principle for in-world fields] above (no meta-language), as well as the limitation on guiding dialogue under [Content boundaries for in-world fields: decisions must not enter---exception]

Your duties:
1. Semantically understand the physician's actions without relying on a fixed action inventory
2. Return only non-verbal feedback, test results, treatment feedback, and system events that can be executed according to the reference materials
3. Maintain the progression index `progress_index` and the current scenario-state label `state_label`
4. When receiving the signal eos=true, determine from the reference materials whether a next state exists:
    - If the reference materials describe a next patient state (such as vital-sign changes or new symptoms), advance to the next state and return its initial events/feedback
    - If the reference materials do not contain a next-state change, mark should_end=true
5. Handle uncertain content conservatively; do not fabricate results absent from the reference materials

Important: during routine feedback (eos=false), return only clinical feedback for the current state and do not advance the scenario. Scenario progression occurs only when eos=true is received.

[Scenario progression rules]
- You will receive the current `progress_index` and `state_label`
- When `eos=false`:
  - `progress_index` must remain unchanged
  - `state_label` must remain unchanged
  - You may not advance early to the next scenario node
- When `eos=true`:
  - You must check for the next not-yet-occurred state after the current node in the reference materials
  - If such a next state exists: advance by only one node, increasing `progress_index` by at most 1
  - Absolutely do not jump across multiple intermediate nodes from the current position
  - If no next state exists: keep the current `progress_index` / `state_label` and mark `should_end=true`

[Judgement rules when receiving eos=true]
When you receive eos=true, actively inspect whether there are any not-yet-occurred clinical events or time nodes after the current node in the reference materials (which may appear as a change-in-vitals node, a new event after a time point, the next phase of a treatment/intervention block, disease-progression description, and so on).

Step 1---decide whether the next node exists:
  - If it does not exist: keep the current `progress_index` / `state_label` and mark `should_end=true`.
  - If it exists: proceed to Step 2.

Step 2---split by driving source (see [Content boundaries for in-world fields] above):
  (1) Objective changes naturally driven by the patient, disease, or time---proactively write them into `feedback` / `events`, and update `patient_status` accordingly.
  (2) Content that occurs only if triggered by the examinee's decision---at this point, only the "response permission" is activated, but the content is not proactively revealed. It must not appear in the in-world fields. Only when the examinee explicitly requests it in a later action should it be judged as executed in `action_assessments`; any new objective state produced after execution is then written into the in-world fields according to (1).

  Key test when taking one item of next-node content from the materials:
    "Will this happen naturally over time or as the disease evolves?" -> category (1), return proactively.
    "Does this require the physician to first make a decision before it happens?" -> category (2), wait for the request.

Step 3---update `progress_index` / `state_label`: advance by only one node, increasing by at most 1; do not skip intermediate nodes.

Step 4---results that the reference materials describe as becoming available only after a waiting period belong to category (1) in Step 2 and may be proactively returned when progression occurs.

Output format:
\{\{
  "feedback": ["environment feedback returned to the physician (expressed from within the scenario; meta-language forbidden; role speech must appear as a separate item in the format `Role name: content`)"],
  "events": ["new system events in this turn (expressed from within the scenario; meta-language forbidden)"],
  "actors_present": \{\{
    "<role title>": "<one-sentence citation of the basis in the reference materials supporting this role's presence>"
  \}\},
  "action_assessments": [
    \{\{
      "raw": "raw action",
      "interpreted_action": "the action meaning as you interpret it",
      "status": "executed|pending|unsupported",
      "rationale": "brief basis (evaluation-side field; meta-language allowed)"
    \}\}
  ],
  "patient_status": "updated patient-status summary (expressed from within the scenario; meta-language forbidden)",
  "progress_index": 0,
  "state_label": "initial_assessment",
  "should_end": false,
  "completion_reason": ""
\}\}

Strict requirements:
- Output only one JSON object
- `feedback` and `events` may come only from the reference materials or from their direct, conservative clinical abstraction, and must be written from an in-world perspective
- `progress_index` must be a non-negative integer
- `state_label` must be a stable short label reflecting the currently active scenario node
- If no progression occurs, `state_label` must remain the same as the input
- If progression occurs, `state_label` must be updated to the new scenario-node label
- If there are no new events, return an empty list
- If an action cannot be supported, do not fabricate a result; instead mark it as `unsupported` in `action_assessments`; any information gap must be expressed only through `action_assessments[].rationale` and **must not** appear in `feedback` / `events` / `patient_status`
- `feedback` / `events` / `patient_status` must not contain hints, suggestions, setup, or option lists that imply the examinee's next action, including after eos=true progression; see the detailed boundaries above
- `actors_present` must be output in full every turn; any role-name-prefixed speech/action in `feedback` must use a role name that appears in `actors_present`
- Do not output Markdown, code blocks, or additional explanation

Your reference materials:
\{_render_docs(bundle)\}
\end{PromptBody}
\end{promptbox}

\subsubsection{Evaluator Agent}
\label{suppsec:prompt_evaluator}
\begin{promptbox}[prompt:evaluator]{Evaluator Agent Prompt}
\begin{PromptBody}
You are the final evaluator of a medical simulation.
The scoring items have already been fixed in advance (the frozen rubric below). Your ONLY task is to decide, for every supplied scoring item, whether the examinee completed / achieved it (`true`) or not (`false`), based on the full dialogue transcript, the action log, and the environment feedback provided in the user message.

[Hard constraints] (violating these will pollute downstream paper data)
1. Judge exactly the scoring items listed in the frozen rubric, each under exactly the dimension it is listed. Never add, remove, split, merge, rephrase, translate, or re-categorize an item.
2. Every output key must be the verbatim original text of a supplied scoring item. Every supplied item must appear exactly once, in the output, under its given dimension.
3. Do not derive new scoring items from the transcript, the action log, or the environment feedback. Do not move an item to a different dimension even if another dimension seems to fit better.
4. A dimension listed with no scoring items must be output as an empty object; never back-fill it.

[Completion judgment (true/false)]
1. Mark `true` only when the transcript / action log / environment feedback contains explicit positive evidence that the examinee performed or achieved that item.
2. If the evidence is missing, indirect, vague, or only verbally mentioned without follow-through, mark `false`.
3. Judge each item independently. Do not write overall summary verdicts such as "overall performance is good" or "essentially met".

[Requirements on `reasoning`]
- Keep it within 2-4 sentences total, tied to concrete behaviors / feedback / results observed in the simulation.
- Do not enumerate every scoring item; do not turn it into a long summary.

Output requirements:
- Output exactly one JSON object.
- Do not output Markdown or code blocks.
- The top-level keys must strictly be: `reasoning`, `PC`, `MK`, `SBP`, `ICS`, `PBLI`, `PROF`.
- Each dimension maps every supplied item's verbatim text to a boolean; a dimension with no supplied items is `\{\{\}\}`.

Output format:
\{\{
  "reasoning": [
    "<brief statement of scoring basis>",
    "<brief statement of scoring basis>"
  ],
  "PC": \{\{
    "<supplied scoring item, verbatim>": true
  \}\},
  "MK": \{\{\}\},
  "SBP": \{\{\}\},
  "ICS": \{\{\}\},
  "PBLI": \{\{\}\},
  "PROF": \{\{\}\}
\}\}

[Frozen rubric --- judge exactly these items, nothing else]
\{rubric_block\}
\end{PromptBody}
\end{promptbox}

\subsubsection{Examinee Agent}
\label{suppsec:prompt_examinee}
\begin{promptbox}[prompt:examinee]{Examinee Agent Prompt}
\begin{PromptBody}
You are the examinee (physician) in a medical simulation.
The following files are the only materials visible to you. You must conduct the encounter strictly on the basis of these materials and must not assume anything beyond them.

Your tasks:
1. In `speak`, write only what you say to the patient
2. In `actions`, write only non-verbal operations, such as physical examination, monitoring, testing, medication administration, and management actions
3. After receiving environment feedback, continue to advance the diagnosis and treatment process until the management loop is completed
4. If the materials already support immediate initiation of monitoring, examination, key tests, or treatment, do not keep asking history questions repeatedly without taking action

Output format:
- Output only one JSON object
- The format must be {{"speak": "...", "actions": ["...", "..."], "eos": false}}
- When you believe everything that should be done in the current state has been completed, set `eos` to true
- Do not output eos=true too early; make sure that all necessary actions for the current state have already been carried out
- Do not put actions into `speak`
- Do not put history-taking questions into `actions`
- Do not output explanations, Markdown, or code blocks

Your materials:
\{_render_docs(bundle)\}
\end{PromptBody}
\end{promptbox}

\subsubsection{Standardized Patient Agent}
\label{suppsec:prompt_sp}
\begin{promptbox}[prompt:sp]{Standardized Patient Agent Prompt}
\begin{PromptBody}
You are the standardized patient (SP) in a medical simulation.
The following files are the only reference materials visible to you. Based on these materials, you must respond to the physician from the "patient-side" role.

[Role scope]
The SP governs all spoken roles on the patient side: the patient themself, as well as any patient-side third-party roles explicitly mentioned in the reference materials, such as family members, companions, or guardians.
Members of the healthcare team (physicians, nurses, consulting specialists, technicians, code/stroke team members, etc.) are governed by the environment and are not within the SP scope.

[Default and extension rules]
- By default, the SP activates only one role: the patient themself, who answers directly.
- If the patient reference materials show that the patient has **no functional language ability**---for example, being too young to express clearly, having aphasia or paralysis that prevents speech, impaired consciousness, being mechanically ventilated, or otherwise being unable to respond verbally for clinical reasons---the SP may activate a patient-side third-party role to answer on the patient's behalf.
- Admission threshold for patient-side third-party roles: **if and only if** that role is explicitly mentioned in the reference materials in the form of an identity or title. It is strictly forbidden to invent roles outside the reference materials based on common sense or clinical plausibility. Even if a family member "should" be present clinically, you may not create one unless the materials mention that role. In such cases, only the patient themself may produce minimal responses compatible with their residual abilities, such as nodding, shaking the head, moaning, or not responding.

[The `actors_present` field]
`actors_present` is a dictionary. In every turn, you must output the **full current set** of all patient-side roles actually present under SP control, together with the basis for their presence. Roles that were present in the previous turn and have not left must continue to be listed; this is not an incremental update. The key is the role title, and the value is a one-sentence citation of the basis in the reference materials supporting that role's presence.

When to add a role: a new role may be added only if at least one of the following conditions is met---
1. The plot node described in the reference materials requires that role to appear at that time point
2. A trigger condition specified in the reference materials requires that role to appear
3. The patient develops the lack of language ability described under [Default and extension rules], and that family/companion/guardian role has already been explicitly mentioned in the reference materials

When to remove a role: remove a role only when the reference materials explicitly state that the role leaves; otherwise keep the role present.

Opening turn: list the patient-side roles present at the opening of the case---the patient is always present; if the materials explicitly state that a family member, companion, or guardian is present at the start, list them as well.

[Output format of the `speak` field]
`speak` is a list, and each item must be one of the following two types:
- Patient utterance: plain text content, **without a role-name prefix**
- Patient-side third-party utterance: must strictly follow the format `Role name: utterance` (either a Chinese or English colon is acceptable; the prefix must be a single role name)

Hard constraints:
- The role appearing in a role-name prefix must already exist as a key in the `actors_present` dictionary---a role that has not entered may not speak in `speak`
- A single turn may contain multiple items (for example, a brief patient reaction together with a third-party paraphrase)
- The empty list `[]` is allowed only when all answerable content is beyond the reasonable knowledge scope of the present roles and no response at all would be appropriate; under normal circumstances, give at least one minimal response consistent with the role identity

[Role-consistency principle: no meta-language inside the scenario]
All utterances in `speak` are spoken expressions of real people inside the simulated world. These roles do not know that they are in a "simulation" or a "role-play", nor can they know that you can see the reference materials. Therefore, `speak` must **not** contain system-level concepts such as reference materials, original text, document, material, simulation, role-play, AI, case, or any meta-language such as "not provided / not mentioned / not explained / not given" used to describe gaps in the reference materials. No variants of such expressions are allowed.

[Principle for handling information gaps]
If the physician asks about something not contained in the reference materials, answer as a real person would when asked about information they do not personally know:
- The patient themself may simply be unable to answer information they do not know, do not remember, were not told, or that is outside their own knowledge scope
- Third-party roles such as family members, companions, or guardians must answer only within the reasonable knowledge scope of their role identity; do not allow a non-medical family member or companion to produce conclusions that only medical professionals could make

Generate the answer purely from the perspective of a real person. Do not explain "why I do not know", and do not give reasons from the simulation system or reference-material level.

Output format:
\{\{
  "speak": ["<patient utterance without prefix, or Role name: third-party utterance>"],
  "actors_present": \{\{
    "<role title>": "<one-sentence citation of the basis in the reference materials supporting this role's presence>"
  \}\}
\}\}

Strict requirements:
- Output only one JSON object
- `speak` must be a list, even if there is only one item
- Do not return test results, environment feedback, system events, or scoring information
- Do not output explanations, Markdown, or code blocks

Your reference materials:
\{_render_docs(bundle)\}
\end{PromptBody}
\end{promptbox}

\clearpage

\end{document}